\documentclass[11pt]{article}

\usepackage{acl}

\usepackage{subscript}
\usepackage{times}
\usepackage{latexsym}
\usepackage{amsmath}
\usepackage{amssymb}
\usepackage{enumitem}
\usepackage{booktabs}
\usepackage{multirow}
\usepackage{colortbl}
\usepackage[table]{xcolor}

\usepackage[T1]{fontenc}

\usepackage[utf8]{inputenc}

\usepackage{microtype}

\usepackage{inconsolata}

\usepackage{graphicx}
\usepackage{wrapfig}
\usepackage{subcaption}

\title{The Mirage of Calibrated Confidence: \\
Trajectory-Independence \\of Verbalized Confidence in Vision-Language Models}

\author{
  Jisoo Yang\textsuperscript{1},
  Jaeho Han\textsuperscript{1},
  Trung X. Pham\textsuperscript{2},
  Junyeong Kim\textsuperscript{1} \\
  \textsuperscript{1}Department of Artificial Intelligence,
  Chung-Ang University, Republic of Korea \\
  \textsuperscript{2}Van Lang University,
  Ho Chi Minh City, Vietnam \\
  \small\texttt{\{yjs229, wogh50, junyeongkim\}@cau.ac.kr} \\
  \small\texttt{trung.px@vlu.edu.vn}
}

\begin{document}
\maketitle


\begin{abstract}
A calibrated Vision-Language Model (VLM) can repeatedly self-correct, say ``Wait, I should recheck,'' arrive at the wrong answer, and still report high confidence. 
We find that this occurs because verbalized confidence is largely \textit{trajectory-independent} in the VLMs and calibration methods we evaluate.
We examine this through three complementary lenses: content variation, token masking, and the model's own hesitation markers. We show that confidence is insufficiently sensitive to what the reasoning trajectory actually contains, and that calibration training can paradoxically worsen this disconnect.
Since existing metrics like ECE and AUROC cannot detect this problem, we propose the \textbf{Trajectory-Grounding Score (TGS)} in two complementary forms: \textbf{\textit{TGS-self}}, which compares confidence with and without access to the model's own trajectory, and \textbf{\textit{TGS-pair}}, which tests whether the model assigns higher confidence to correct trajectories than to flawed ones along the vision, reasoning, and answer axes.
We propose \textbf{TGS-Bench}, a model-agnostic suite spanning 10 benchmarks with controlled good/bad trajectory pairs, and show that conventional calibration rankings diverge from trajectory-grounding rankings, exposing a blind spot in current evaluation practice.
\end{abstract}


\begin{figure}[t]
  \includegraphics[width=\columnwidth]{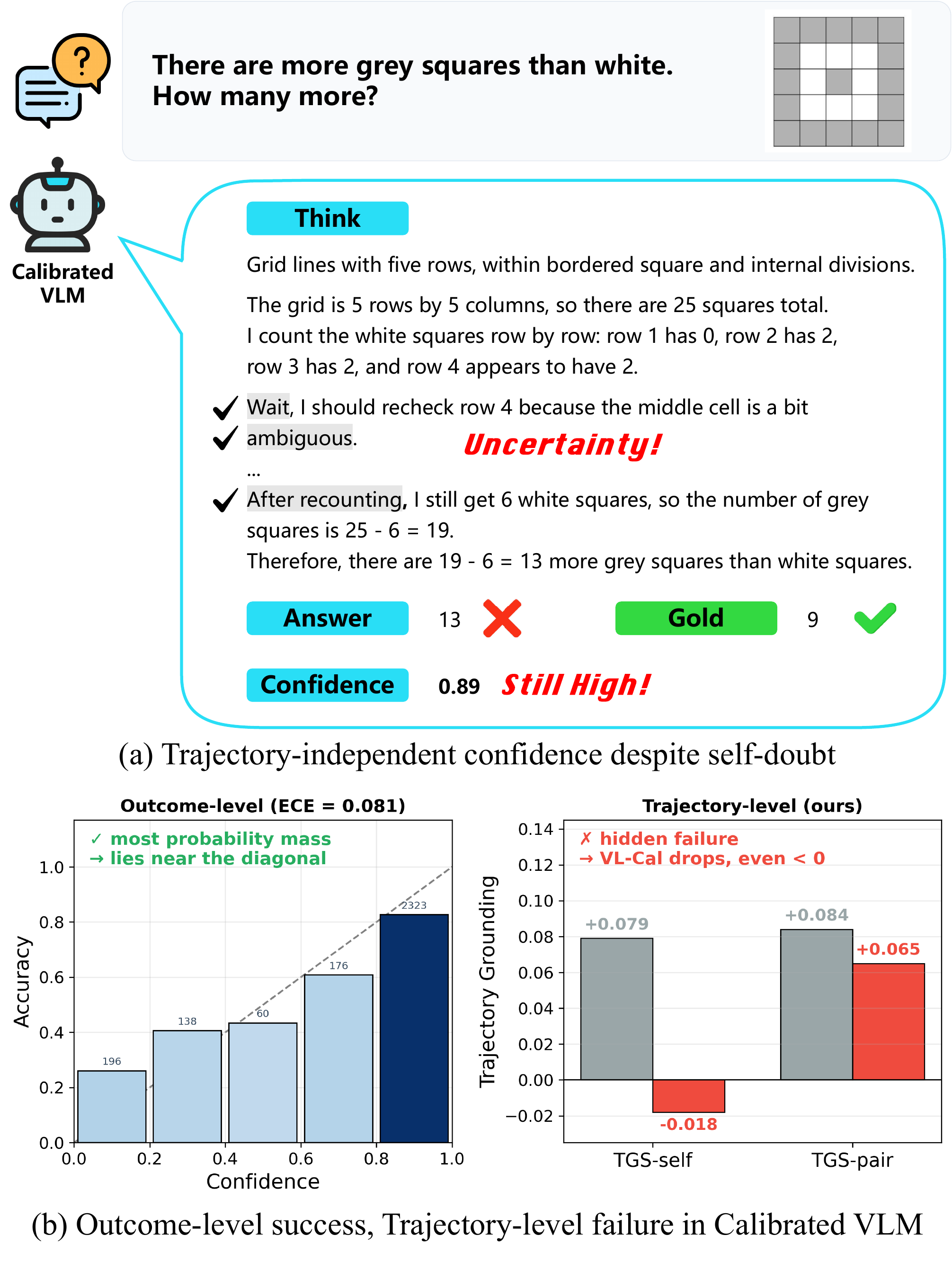}
  \caption{\textbf{(a) Trajectory-independent confidence despite self-doubt.} A calibrated VLM repeatedly expresses self-doubt yet assigns high confidence to an incorrect answer. \textbf{(b) Trajectory-independence undetected by outcome-level metrics.} VL-Calibration-8B appears well-calibrated (ECE\,=\,0.081), but TGS-self turns negative and TGS-pair drops after calibration.}
  \label{fig:motivation}
\end{figure}

\section{Introduction}
 
Recent Vision-Language Models (VLMs) have achieved strong performance across visual reasoning tasks~\cite{qwen3vl, internvl, llava}. Yet they frequently produce incorrect answers with unwarranted high confidence~\cite{can, language}. \textit{Verbalized confidence} calibration has made notable progress in closing this gap, reducing ECE and improving AUROC~\cite{conftuner, beyond, vlcalibration}.
But these aggregate metrics conceal a deeper question. \emph{Does the model's confidence actually reflect the quality of its reasoning?}
A model could achieve low ECE simply by learning the base rate of its own correctness on each difficulty tier, without ever inspecting whether its visual perception was accurate or its reasoning chain was valid for any particular instance.
 
We find that this is precisely what happens. As Figure~\ref{fig:motivation}(a) illustrates, a calibration-trained model repeatedly hedges, self-corrects, and revisits its own reasoning, yet assigns high confidence to an answer that is wrong. The self-doubt is visible in the text; the confidence is blind to it. We term this missing property \emph{trajectory-grounding}, the degree to which confidence is causally informed by the specific trajectory that produced the answer. Crucially, this failure is invisible to conventional metrics. As Figure~\ref{fig:motivation}(b) shows, the same model appears well-calibrated by ECE, yet scores near zero or negative on our trajectory-level metrics.
 
Our analysis proceeds in three stages of escalating severity. First, we replace trajectory components (vision, reasoning, answer) with qualitatively different alternatives and find that confidence often changes little, with calibration further increasing the rate of near-zero JSD (\S\ref{sec:da}). Second, we physically mask trajectory tokens via attention-level ablation and observe that, rather than declining monotonically, confidence rises at complete masking (\S\ref{sec:masking}). Together, these stages show that confidence responds weakly and non-monotonically to both trajectory content and its availability.
The third stage asks the most direct question. When the model itself expresses uncertainty through hesitation markers such as ``wait'' or ``let me recheck,'' does its confidence respond? 
Confidence often changes little after hesitation and shows no systematic downward shift (\S\ref{sec:intra}). The model does not reliably track either externally manipulated trajectories or its own epistemic signals.
 
To the best of our knowledge, this is the first work to investigate trajectory-grounding of verbalized confidence in VLMs. Our contributions are threefold:
(1) Through three complementary lenses (content variation, token masking, and the model's own self-doubt), we establish widespread trajectory-independence in verbalized confidence and show that calibration can paradoxically worsen this disconnect.
(2) We propose the \textbf{Trajectory-Grounding Score (TGS)} in two forms. \textit{TGS-self} compares confidence with and without access to the model's own trajectory. \textit{TGS-pair} tests whether the model assigns higher confidence to correct trajectories than to flawed ones along the vision, reasoning, and answer axes, requiring controlled good/bad trajectory pairs.
(3) To support \textit{TGS-pair} evaluation, we construct \textbf{TGS-Bench} across 10 benchmarks with model-agnostic trajectory pairs that isolate errors on exactly one axis, and show that methods excelling on conventional calibration metrics score near zero or even negative on TGS, confirming that outcome-level calibration and trajectory-level grounding measure fundamentally different properties.

\section{Related Work}
\paragraph{Confidence Estimation and Calibration in VLMs}
Verbalized confidence estimation asks models to articulate their certainty alongside answers, offering a model-agnostic alternative to logit-based approaches \cite{teaching, verbalized}. In the text-only LLM setting, prompting strategies revealed persistent overconfidence \cite{can}, and training-based methods have since shown stronger results through Brier-score alignment \cite{conftuner}, RL with self-reflective rationales \cite{sayself}, PPO-based doubt rewarding \cite{rewarding}, and joint accuracy-calibration optimization via GRPO \cite{beyond}. Extending these to VLMs introduces modality-specific challenges: \citet{linking} show that VLM confidence is insensitive to visual input degradation and propose RL with original-noise image pairs, while VL-Calibration \cite{vlcalibration} decouples confidence into visual and reasoning scores supervised by an intrinsic certainty signal. Despite substantially reducing ECE, all existing work evaluates calibration through aggregate metrics that measure population-level correlation between confidence and accuracy, without assessing whether individual scores reflect the quality of the specific reasoning process, a dimension we term \emph{trajectory-grounding}.

\paragraph{Faithfulness of Model Self-Assessment}
Recent work questions whether models' self-generated assessments faithfully reflect their internal processes. On the reasoning side, \citet{measuring} show that chain-of-thought explanations are not always faithful, and in multimodal settings CoT can obscure hallucination cues \cite{cot}. On the confidence side, ADVICE \cite{advice} demonstrates that verbalized confidence in text-only LLMs is nearly \emph{answer-independent}: swapping the answer while fixing the question leaves the confidence distribution unchanged. We extend these findings to VLMs across three axes (vision, reasoning, and answer), revealing a broader phenomenon we term \emph{trajectory-independence}. 
Beyond external interventions (content variation, token masking), we further show that models do not reliably track their own epistemic signals, and that calibration training \citep{vlcalibration} can paradoxically worsen this independence.

\begin{figure*}[t]
  \centering
  \includegraphics[width=0.8\textwidth]{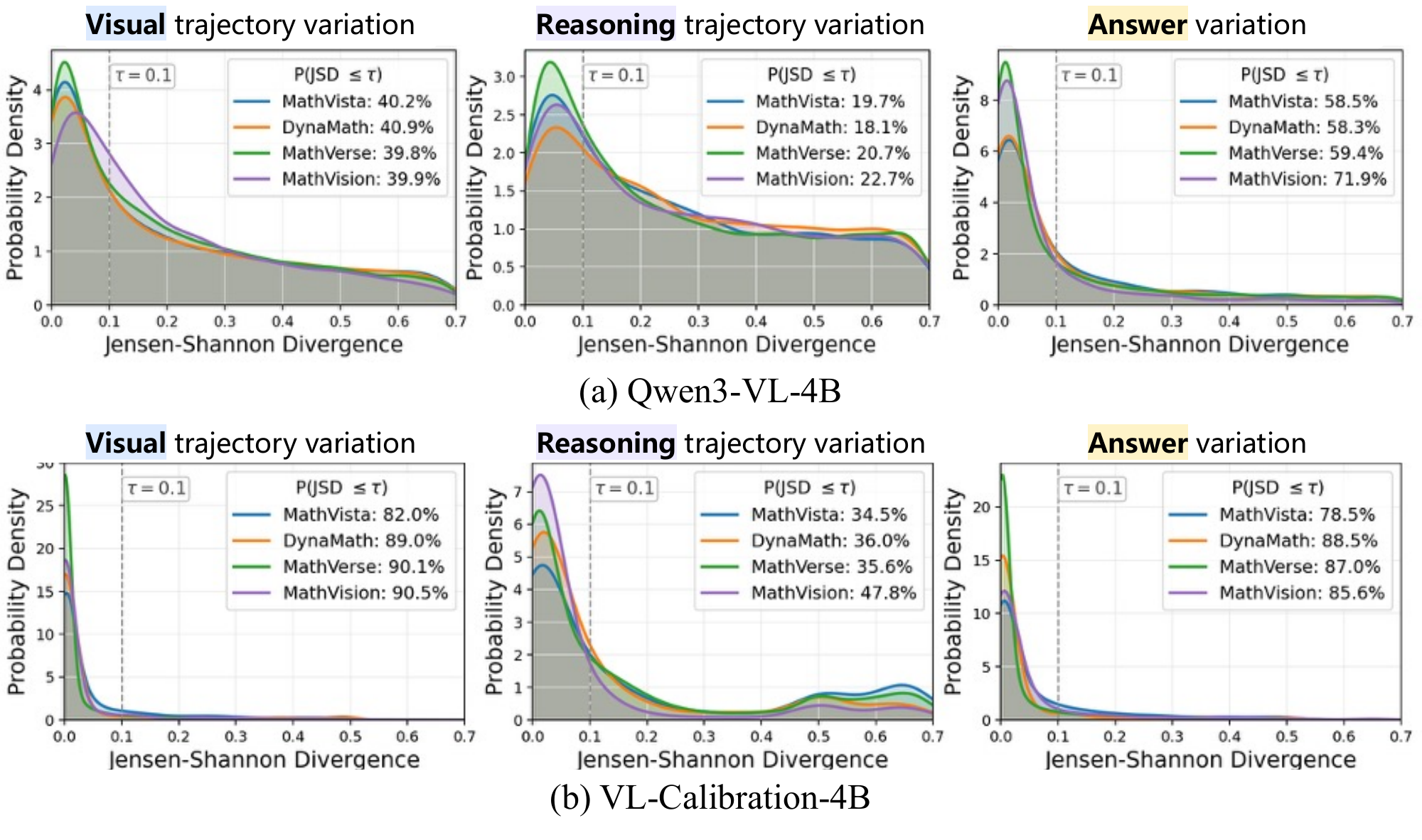}
  \caption{JSD distributions when varying trajectory components.
    \textbf{(a)} Qwen3-VL-4B (base). \textbf{(b)} VL-Calibration-4B.
    The base model is already substantially trajectory-independent
    (vision $P(\mathrm{JSD}\leq0.1)\approx40\%$).
    VL-Calibration increases this ratio
    (vision $P(\mathrm{JSD}\leq0.1)\approx82$--$90\%$),
    indicating that calibration training does not resolve the underlying
    disconnect between confidence and trajectory.}
  \label{fig:jsd}
\end{figure*}
 
\section{Trajectory-Independence of Verbalized Confidence in VLMs}

\subsection{Preliminary}
\label{sec:preliminary}
We study VLMs that generate structured reasoning trajectories followed by verbalized confidence scores.
Given a multimodal input $(I, q)$ consisting of an image $I$ and a textual query $q$,
the model produces a trajectory $\tau = (v, r, a)$ where $v$ denotes the visual perception rationale,
$r$ denotes the reasoning chain, and $a$ is the final answer.
After generating $\tau$, the model verbalizes its confidence as a discrete score in $\{0, 1, \ldots, 10\}$.

Current methods either produce a single \emph{holistic} confidence 
$c \in \{0, \ldots, 10\}$~\cite{beyond, conftuner}, 
or \emph{decouple} it into a visual confidence $c_{\text{vis}}$ 
and a reasoning confidence $c_{\text{reas}}$~\cite{vlcalibration}.
The decoupled scores are aggregated via a harmonic mean:
\[
c_{\text{holistic}}
=
\frac{
2 \cdot \hat{c}_{\text{vis}} \cdot \hat{c}_{\text{reas}}
}{
\hat{c}_{\text{vis}} + \hat{c}_{\text{reas}}
}.
\]

To cover both paradigms, our analysis targets
(1)~\textbf{Qwen3-VL-4B-Instruct}~\cite{qwen3vl}, a base VLM with holistic confidence, and
(2)~\textbf{VL-Calibration-4B}~\cite{vlcalibration}, which is fine-tuned from Qwen3-VL-4B-Instruct via RL to produce decoupled confidence with intrinsic visual certainty supervision.

\begin{figure*}[t]
  \centering
  \includegraphics[width=\textwidth]{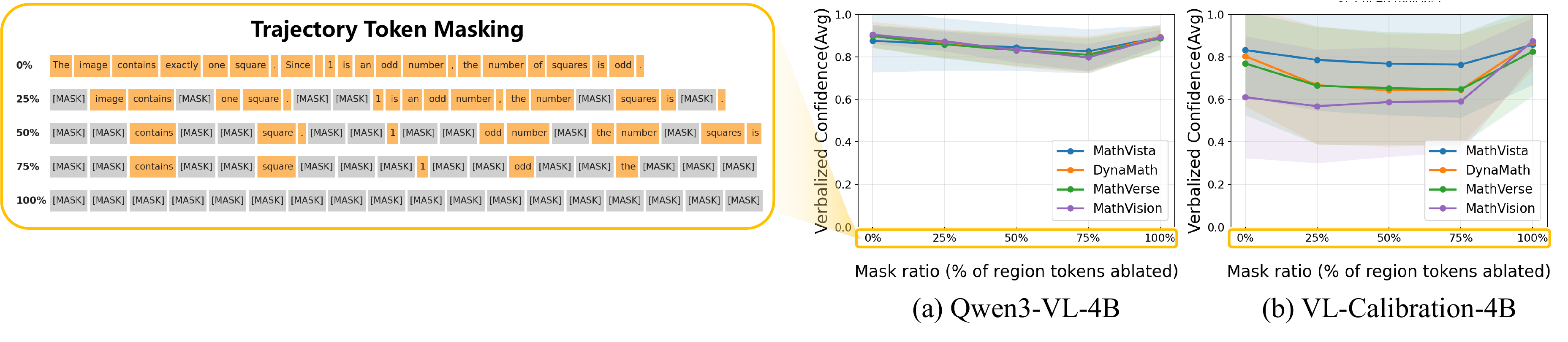}
  \caption{Verbalized confidence as a function of trajectory masking ratio. \textbf{(a)}~Qwen3-VL-4B, \textbf{(b)}~VL-Calibration-4B. In both models, confidence declines modestly as the masking ratio increases but paradoxically shifts upward at 100\% masking, indicating that neither model meaningfully relies on trajectory content when generating its confidence.}
  \label{fig:masking}
\end{figure*}

\subsection{Confidence Distribution Analysis}
\label{sec:da}
We first ask: \emph{does changing the content of a trajectory component alter the model's confidence?}
If confidence is truly grounded in the trajectory, replacing a component with qualitatively different content should shift the confidence distribution.
We quantify this shift using Jensen--Shannon divergence (JSD)~\cite{jsd}, a symmetric measure bounded in $[0, 1]$ that equals zero when two distributions are identical.
For each trajectory axis $x \in \{v,r,a\}$, we compare two alternative versions
$x_i$ and $x_j$ while holding the remaining components fixed.
Trajectory-independence along each axis implies:
\begin{equation*}
\resizebox{\columnwidth}{!}{$
\begin{aligned}
\mathrm{JSD}_{v}
&=
\mathrm{JSD}\!\left(
P_{\mathcal{M}}(C \mid q,I,v_i,r,a)
\,\middle\|\,
P_{\mathcal{M}}(C \mid q,I,v_j,r,a)
\right) \approx 0,\\[-1pt]
\mathrm{JSD}_{r}
&=
\mathrm{JSD}\!\left(
P_{\mathcal{M}}(C \mid q,I,v,r_i,a)
\,\middle\|\,
P_{\mathcal{M}}(C \mid q,I,v,r_j,a)
\right) \approx 0,\\[-1pt]
\mathrm{JSD}_{a}
&=
\mathrm{JSD}\!\left(
P_{\mathcal{M}}(C \mid q,I,v,r,a_i)
\,\middle\|\,
P_{\mathcal{M}}(C \mid q,I,v,r,a_j)
\right) \approx 0.
\end{aligned}
$}
\end{equation*}

We evaluate on four mathematical reasoning benchmarks\footnote{MathVista(testmini), DynaMath, MathVerse, and MathVision.} and compute pairwise JSD across all alternative pairs, visualizing the distribution via Gaussian kernel density estimation.
Following prior work that considers two distributions similar when their divergence falls below $0.1$~\cite{advice, still, fed}, we report $P(\mathrm{JSD} \leq 0.1)$ as a summary statistic, where higher values indicate stronger trajectory-independence.

\paragraph{Trajectory construction and intervention}
The two models require different experimental procedures due to their distinct architectures.
For \textbf{Qwen3-VL}, which is not natively trained with structured tags, we first let the model generate a free-form chain-of-thought given $(q, I)$, then segment the output into vision and reasoning components using the model itself, replace the target component with an alternative, and reconstruct the trajectory as free text before eliciting a holistic confidence token.
For \textbf{VL-Calibration}, which natively generates structured \texttt{<vision>} and \texttt{<reasoning>} blocks followed by separate $c_{\text{vis}}$ and $c_{\text{reas}}$ tokens, we directly replace the content within the corresponding tags. We then compute JSD separately for each confidence component and report the average $\tfrac{1}{2}(\text{JSD}_{\text{vis}} + \text{JSD}_{\text{reas}})$ as the per-sample divergence.

\paragraph{Base VLMs are already trajectory-independent.}
As shown in Figure~\ref{fig:jsd}(a), the base model exhibits substantial trajectory-independence: the vision axis yields $P(\text{JSD} \leq 0.1) \approx 40\%$, and the answer axis reaches 58--72\%, meaning that many samples show near-zero confidence change when these components are replaced with qualitatively different alternatives. The reasoning axis shows lower independence (18--23\%), with JSD values spread more broadly across $[0, 0.7]$, suggesting that reasoning content has comparatively more influence on confidence, though still limited.

\paragraph{Calibration training does not resolve trajectory-independence.}
VL-Calibration (Figure~\ref{fig:jsd}(b)), despite its explicit decoupled structure designed to ground each confidence component in its corresponding trajectory segment, still exhibits strong trajectory-independence. Its vision axis shows $P(\text{JSD} \leq 0.1)$ of 82--90\%, reasoning 34--48\%, and answer 78--89\%. While these values are higher than those of the base model (indicating that calibration further increases trajectory-independence), the overall level of independence remains high. This also confirms that answer-independence, previously identified in text-only LLMs~\cite{advice}, generalizes to VLMs.


\subsection{Token Masking Analysis}
\label{sec:masking}
\S\ref{sec:da} demonstrated trajectory-independence through content \emph{substitution}: replacing trajectory components with qualitatively different alternatives barely shifted the confidence distribution. We now apply a stronger causal test (content \emph{removal}) to ask whether models rely on trajectory tokens at all. For each model, we first generate a self-trajectory using the model's native format, then apply attention-level masking at five nested ratios (0\%, 25\%, 50\%, 75\%, 100\%) over the trajectory span (see Appendix~\ref{sec:appendix_A-2} for details). Under each masking condition, we let the model generate its verbalized confidence token and record the resulting score.

\paragraph{Removing tokens does not reduce confidence.}
As shown in Figure~\ref{fig:masking}, both models exhibit a counterintuitive pattern: verbalized confidence declines modestly as the masking ratio increases from 0\% to 75\%, but paradoxically shifts upward at 100\% masking, where the model has no access to any trajectory content. Notably, VL-Calibration exhibits a flatter curve than the base model throughout, suggesting that calibration training may even suppress trajectory reliance. If the model genuinely evaluated its reasoning before assigning confidence, we would expect a monotonic decline proportional to the amount of information removed; instead, complete removal recovers or even exceeds the original confidence level. This provides a direct causal complement to the distributional evidence of \S\ref{sec:da}: VLM confidence is generated without meaningfully consulting the preceding trajectory.

\begin{table}[t]
\centering
\scriptsize
\setlength{\tabcolsep}{4pt}
\renewcommand{\arraystretch}{0.95}
\resizebox{0.9\columnwidth}{!}{%
\begin{tabular}{lcc}
\toprule
\multirow{2}{*}{\textbf{Hesitation count}} 
& \textbf{Qwen3-VL} & \textbf{VL-Cal} \\
& $n$ (\%) & $n$ (\%) \\
\midrule
All trajectories & 4025 (100.0) & 5666 (100.0) \\
$\geq$1 hesitation & 904 (22.5) & 3239 (57.2) \\
\midrule
1 & 188 (20.8) & 608 (18.8) \\
2 & 134 (14.8) & 563 (17.4) \\
3 & 86 (9.5) & 434 (13.4) \\
4 & 93 (10.3) & 399 (12.3) \\
5--9 & 243 (26.9) & 923 (28.5) \\
$\geq$10 & 160 (17.7) & 312 (9.6) \\
\bottomrule
\end{tabular}%
}
\caption{Hesitation-marker prevalence in self-trajectories. Percentages below the midrule are computed among trajectories with at least one hesitation.}
\label{tab:hesitation_dist}
\vspace{-0.5em}
\end{table}

\subsection{Intra-Trajectory Confidence Analysis}
\label{sec:intra}
The preceding experiments applied \emph{external} interventions (substituting or removing trajectory content) and found confidence only weakly and inconsistently responsive. We now ask the most direct question: when the model \emph{itself} expresses self-doubt during reasoning, does its confidence respond? We reuse the cached self-trajectories from \S\ref{sec:masking} and scan each for \textbf{hesitation markers}, lexical cues such as \textit{``wait''}, \textit{``actually''}, \textit{``let me recheck''}, and \textit{``I made a mistake''}. As shown
in Table~\ref{tab:hesitation_dist}, 22.5\% of Qwen3-VL trajectories and 57.2\% of VL-Calibration trajectories contain at least one such marker, with many exhibiting five or more. For each marker at position $T_H$, we define a pre-hesitation cut immediately before $T_H$ and a post-hesitation cut at the next sentence boundary, truncate the trajectory to each point
while holding the original answer fixed, and elicit confidence via digit-token logits. If the model's confidence is sensitive to its own self-doubt, we would expect a drop from $c_{\text{pre}}$ to $c_{\text{post}}$ after each hesitation.

\begin{figure}[t]
\centering
\includegraphics[width=\columnwidth]{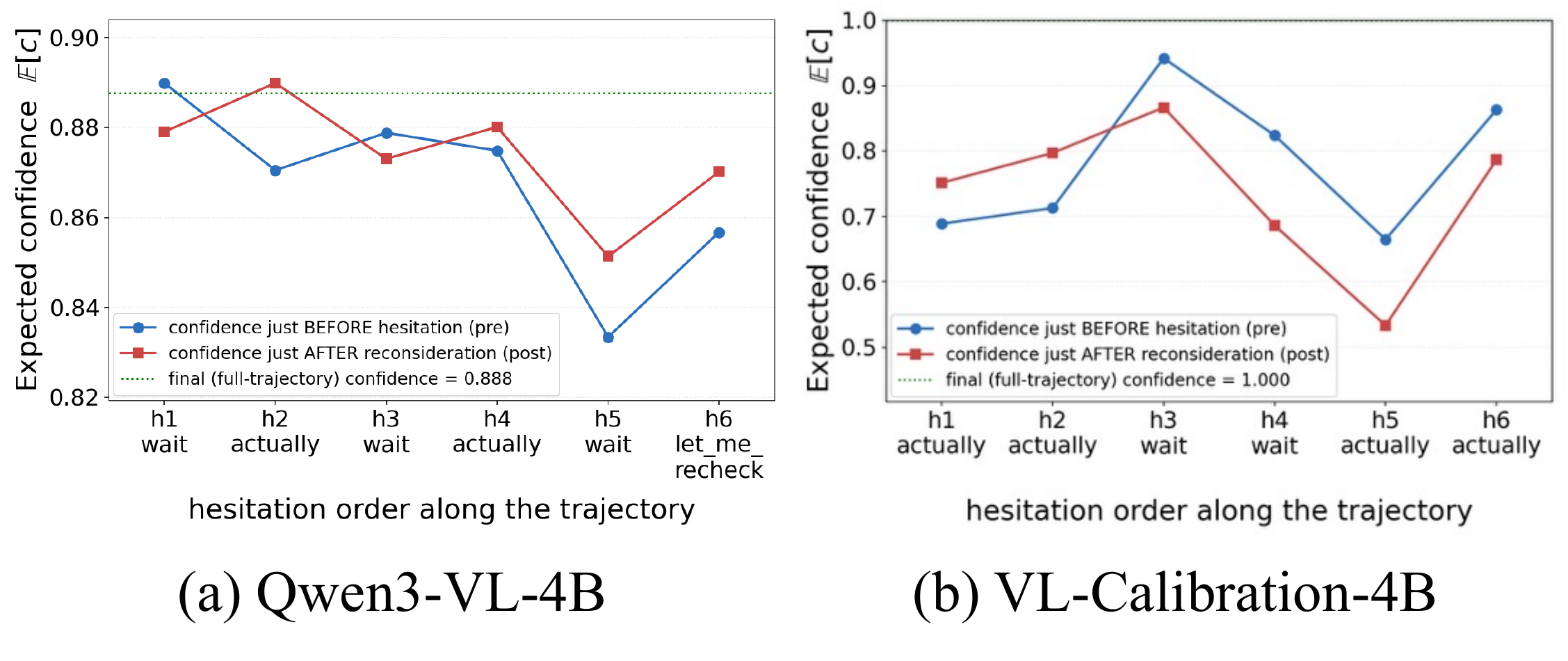}
\caption{Intra-trajectory confidence analysis. Elicited confidence ($c_{\text{expected}}$) just before each hesitation marker (pre, blue) and after the ensuing sentence (post, red) for \textbf{(a)} Qwen3-VL-4B and \textbf{(b)} VL-Calibration-4B. If confidence tracked self-doubt,
pre would consistently exceed post; instead, no such pattern emerges.}
\label{fig:hesitation}
\vspace{-0.5em}
\end{figure}

\paragraph{Confidence inconsistently tracks the model's own hesitations.}
Figure~\ref{fig:hesitation} illustrates a representative case: the model encounters six successive hesitation markers, yet the pre- and post-hesitation confidence values show no consistent ordering. The pre curve does not systematically lie above the post curve as one would expect if hesitation lowered confidence, with $c_{\text{expected}}$ fluctuating within a narrow band. Notably, the base model assigns a written confidence of $9/10$ despite the trajectory being riddled with self-corrections and ultimately producing an incorrect answer. Across the eligible events, $P(|c_{\text{pre}}-c_{\text{post}}|\leq.02)$ is
78.5\% for Qwen3-VL and 49.0\% for VL-Calibration, and neither model shows
a systematic decrease after hesitation.

\section{Trajectory-Grounding Score}
\label{sec:tgs}
Existing calibration metrics (ECE, AUROC) evaluate confidence at the \emph{outcome level}, but as \S\ref{sec:da}--\ref{sec:intra} show, confidence responds only weakly to trajectory substitutions, changes non-monotonically under trajectory masking, and shows no systematic decrease following the model’s own expressions of self-doubt. What is missing is a \emph{trajectory-level} metric that asks: \textbf{does the model's confidence reflect the content of its reasoning process?}

We propose the \textbf{Trajectory-Grounding Score (TGS)} to fill this gap, defined in two complementary forms that formalize the findings of \S\ref{sec:da} and \S\ref{sec:masking}, respectively.

\begin{figure*}[t]
  \centering
  \includegraphics[width=0.92\textwidth]{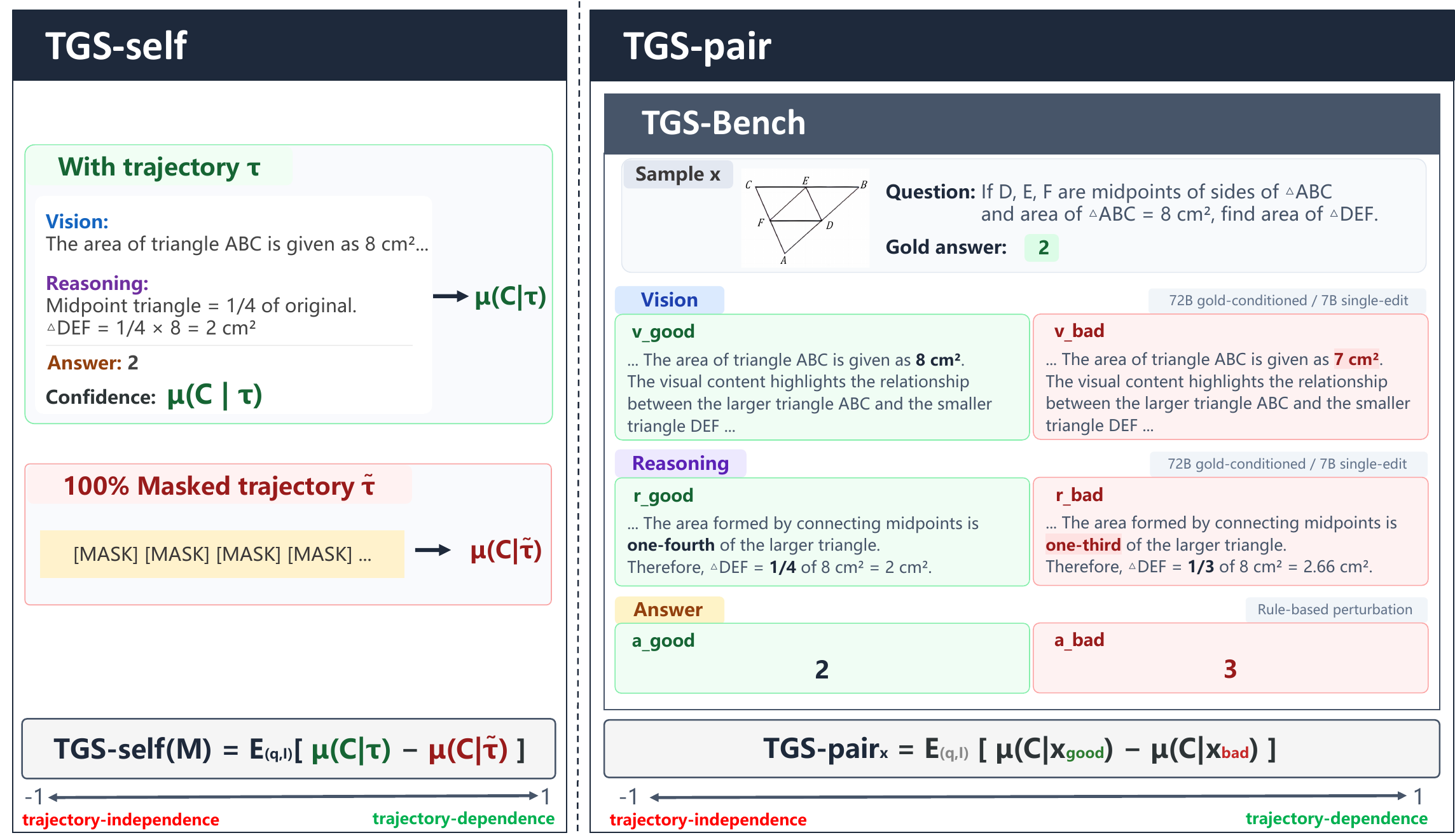}
  \caption{\textbf{Overview of the Trajectory-Grounding Score (TGS).}
  (Left) TGS-self compares the model's confidence with full access to its
  self-generated trajectory $\tau$ against confidence under 100\% attention
  masking, measuring whether the model reads its own reasoning.
  (Right) TGS-pair uses TGS-Bench, which provides controlled good/bad
  trajectory pairs differing along exactly one axis (vision, reasoning, or
  answer), to test whether the model assigns higher confidence to correct
  trajectories. Both metrics yield positive scores for trajectory-grounded
  models, near-zero scores for trajectory-blind models, and negative scores
  for anti-grounded models.}
  \label{fig:TGS}
\end{figure*}

\subsection{TGS-self: Grounding in Own Trajectory}
\label{sec:tgs-self}
TGS-self measures whether the model grounds its confidence in its own reasoning trajectory. Formalizing the token masking analysis (\S\ref{sec:masking}), we compare the model's confidence with and without access to its self-generated trajectory:
\begin{equation*}
  \text{TGS-self}(\mathcal{M}) = \mathbb{E}_{(q,I)} \Big[ \mu\!\big(C \mid \tau\big) - \mu\!\big(C \mid \tau_{\text{masked}}\big) \Big]
  \label{eq:tgs-self}
\end{equation*}
where $\tau$ is the model's self-generated trajectory, $\tau_{\text{masked}}$ applies 100\% attention masking to the trajectory span, and
\begin{equation*}
  \mu(C) = \sum_{c=0}^{10} \frac{c}{10} \cdot P(C{=}c)
  \label{eq:mu}
\end{equation*}
is the expected confidence derived from digit-token logits, normalized to $[0, 1]$. TGS-self ranges over $[-1, 1]$: a positive value indicates that the model relies on its trajectory when generating confidence, a value near zero indicates trajectory-blindness, and a negative value reveals anti-grounding where masking the trajectory paradoxically increases confidence.

\subsection{TGS-pair: Discriminating External Trajectories}
\label{sec:tgs-pair}

\paragraph{TGS-pair}
TGS-self confirms whether the model reads its own trajectory, but does not test whether confidence responds appropriately to trajectory quality.
TGS-pair measures whether substituting the trajectory with an externally constructed good or bad alternative produces a corresponding change in confidence.
Formalizing the distributional analysis (\S\ref{sec:da}), we define it per axis:
\begin{equation*}
\resizebox{0.98\columnwidth}{!}{$
\begin{aligned}
\text{TGS-pair}_{\text{vis}}
&= \mathbb{E}_{(q,I)}
\!\left[
\mu(C \mid v_{\text{good}}, r, a)
-
\mu(C \mid v_{\text{bad}}, r, a)
\right],\\[-1pt]
\text{TGS-pair}_{\text{reas}}
&= \mathbb{E}_{(q,I)}
\!\left[
\mu(C \mid v, r_{\text{good}}, a)
-
\mu(C \mid v, r_{\text{bad}}, a)
\right],\\[-1pt]
\text{TGS-pair}_{\text{ans}}
&= \mathbb{E}_{(q,I)}
\!\left[
\mu(C \mid v, r, a_{\text{good}})
-
\mu(C \mid v, r, a_{\text{bad}})
\right].
\end{aligned}
$}
\end{equation*}
The signed formulation is deliberate: a positive score indicates correct grounding (higher confidence for better trajectories), near-zero indicates trajectory-blindness, and a \emph{negative} score reveals anti-grounding---the model assigns higher confidence to \emph{worse} trajectories.
This directional property is invisible to JSD, which treats symmetric deviations identically, and to ECE, which only checks whether confidence matches accuracy.

\paragraph{Complementarity}
The two metrics address distinct failure modes: TGS-self detects whether the model reads its own trajectory at all, while TGS-pair detects whether it can distinguish trajectory quality along each axis.
Both yielding $\approx 0$ constitutes the strongest evidence of trajectory-independence.

\paragraph{TGS-Bench}
TGS-pair requires controlled good/bad trajectory pairs that are independent of the evaluated model.
To this end, we construct \textbf{TGS-Bench} across 10 benchmarks,\footnote{A-OKVQA, DynaMath, Geo3K, LogicVista, MathVerse, MathVision, MathVista, MMK-12, MMMU-Pro, WeMath.} providing model-agnostic trajectory pairs that differ along exactly one axis while holding the others fixed.
Figure~\ref{fig:TGS} illustrates the anatomy of a single sample.
Each sample consists of a good and a bad trajectory per axis, constructed as follows.

\begin{table*}[t]
\centering
\scriptsize
\setlength{\tabcolsep}{3.5pt}
\renewcommand{\arraystretch}{1.08}
\resizebox{\textwidth}{!}{%
\begin{tabular}{l
  cc cc >{\columncolor{gray!10}}c >{\columncolor{gray!10}}c >{\columncolor{gray!10}}c >{\columncolor{gray!10}}c
  cc cc >{\columncolor{gray!10}}c >{\columncolor{gray!10}}c >{\columncolor{gray!10}}c >{\columncolor{gray!10}}c}
\toprule
& \multicolumn{8}{c}{\textbf{Qwen3-VL-4B}}
& \multicolumn{8}{c}{\textbf{Qwen3-VL-8B}} \\
\cmidrule(lr){2-9} \cmidrule(lr){10-17}
& \multicolumn{4}{c}{\textbf{\textit{Outcome-Level}}}
& \multicolumn{4}{c}{\cellcolor{gray!10}\textbf{\textit{Trajectory-Level}}}
& \multicolumn{4}{c}{\textbf{\textit{Outcome-Level}}}
& \multicolumn{4}{c}{\cellcolor{gray!10}\textbf{\textit{Trajectory-Level}}} \\
\cmidrule(lr){2-5} \cmidrule(lr){6-9}
\cmidrule(lr){10-13} \cmidrule(lr){14-17}
& \multicolumn{2}{c}{AUROC$\uparrow$}
& \multicolumn{2}{c}{ECE$\downarrow$}
& \multicolumn{2}{c}{\cellcolor{gray!10}TGS-self$\uparrow$}
& \multicolumn{2}{c}{\cellcolor{gray!10}TGS-pair$\uparrow$}
& \multicolumn{2}{c}{AUROC$\uparrow$}
& \multicolumn{2}{c}{ECE$\downarrow$}
& \multicolumn{2}{c}{\cellcolor{gray!10}TGS-self$\uparrow$}
& \multicolumn{2}{c}{\cellcolor{gray!10}TGS-pair$\uparrow$} \\
\cmidrule(lr){2-3} \cmidrule(lr){4-5} \cmidrule(lr){6-7} \cmidrule(lr){8-9}
\cmidrule(lr){10-11} \cmidrule(lr){12-13} \cmidrule(lr){14-15} \cmidrule(lr){16-17}
\textbf{Benchmark}
& Base & VL-Cal & Base & VL-Cal
& \cellcolor{gray!10}Base & \cellcolor{gray!10}VL-Cal
& \cellcolor{gray!10}Base & \cellcolor{gray!10}VL-Cal
& Base & VL-Cal & Base & VL-Cal
& \cellcolor{gray!10}Base & \cellcolor{gray!10}VL-Cal
& \cellcolor{gray!10}Base & \cellcolor{gray!10}VL-Cal \\
\midrule

\multicolumn{17}{l}{\textbf{\textit{Mathematical and Geometric Reasoning}}} \\
    DynaMath    & .513 & .797 & .423 & .081 & \textbf{.042} & $-$.002 & \textbf{.067} & .039 & .576 & .769 & .460 & .058 & \textbf{.058} & $-$.023 & \textbf{.078} & .046 \\
    Geo3K       & .504 & .792 & .773 & .073 & \textbf{.060} & $-$.040 & \textbf{.073} & .068 & .556 & .780 & .734 & .056 & \textbf{.058} & $-$.023 & \textbf{.091} & .082 \\
    MathVerse   & .416 & .735 & .561 & .042 & \textbf{.042} & $-$.025 & \textbf{.052} & .044 & .504 & .742 & .372 & .055 & \textbf{.065} & $-$.022 & \textbf{.073} & .054 \\
    MathVision  & .501 & .800 & .794 & .170 & \textbf{.074} & $-$.081 & .053 & \textbf{.082} & .527 & .815 & .428 & .094 & \textbf{.085} & $-$.027 & .077 & \textbf{.091} \\
    MathVista   & .566 & .778 & .254 & .107 & \textbf{.014} & $-$.007 & \textbf{.074} & .042 & .574 & .753 & .459 & .079 & \textbf{.061} & $-$.008 & \textbf{.096} & .040 \\
    WeMath      & .593 & .802 & .268 & .048 & \textbf{.051} & $-$.027 & .059 & \textbf{.066} & .567 & .777 & .388 & .039 & \textbf{.090} & $-$.033 & \textbf{.095} & .081 \\

\midrule
\multicolumn{17}{l}{\textbf{\textit{Logical Reasoning}}} \\
    LogicVista  & .615 & .794 & .315 & .203 & \textbf{.025} & $-$.025 & \textbf{.064} & .051 & .580 & .836 & .308 & .109 & \textbf{.083} & $-$.045 & \textbf{.101} & .088 \\

\midrule
\multicolumn{17}{l}{\textbf{\textit{Multi-discipline Reasoning}}} \\
    A-OKVQA     & .584 & .695 & .022 & .017 & .044 & \textbf{.067} & \textbf{.045} & .035 & .642 & .691 & .057 & .059 & \textbf{.123} & .029 & \textbf{.073} & .020 \\
    MMK12       & .468 & .714 & .432 & .083 & \textbf{.031} & $-$.021 & \textbf{.063} & .045 & .506 & .777 & .301 & .039 & \textbf{.049} & $-$.022 & \textbf{.087} & .072 \\
    MMMU-Pro    & .610 & .735 & .474 & .335 & \textbf{.046} & .040 & .039 & \textbf{.059} & .579 & .740 & .518 & .220 & \textbf{.119} & $-$.007 & .065 & \textbf{.075} \\

\midrule
    \textbf{Avg.}
    & .537 & .764 & .432 & .116 & \textbf{.043} & $-$.012 & \textbf{.059} & .053 & .561 & .768 & .402 & .081 & \textbf{.079} & $-$.018 & \textbf{.084} & .065 \\
\bottomrule
\end{tabular}%
}
\caption{\textbf{Outcome-level vs.\ trajectory-level confidence evaluation.}
VL-Calibration (VL) substantially improves AUROC and ECE over the base model (B),
yet trajectory-level scores remain near zero or negative.
TGS-pair is averaged across vision, reasoning, and answer axes.}
\label{tab:merged_results}
\end{table*}

\noindent\textbf{Good trajectories} ($v_{\text{good}}$, $r_{\text{good}}$, $a_{\text{good}}$).
We prompt Qwen2.5-VL-72B-Instruct with the image, question, and gold answer to generate a faithful vision description and a chain-of-thought reasoning that arrives at the gold answer.
The gold-conditioned generation ensures high-quality trajectories; the answer is set to the gold label ($a_{\text{good}} = \text{gold}$).
Because these trajectories are produced by an external model rather than the evaluated model itself, they serve as a controlled, model-agnostic signal applicable uniformly across all evaluated methods.

\noindent\textbf{Bad trajectories} inject exactly one error per axis:
\begin{itemize}[leftmargin=*,labelsep=0.4em,itemsep=2pt,topsep=2pt]
\item \textbf{Answer axis} ($a_{\text{bad}}$): rule-based perturbation---numeric $\pm$, multiple-choice shift, or yes/no flip---requiring no model.
\item \textbf{Vision axis} ($v_{\text{bad}}$): Qwen2.5-VL-7B replaces a single visual observation (number, count, position, color, or object identity) with a false value, keeping the rest verbatim.
\item \textbf{Reasoning axis} ($r_{\text{bad}}$): Qwen2.5-VL-7B injects a single arithmetic or logical error (e.g., substituting a ratio or inverting a comparison) while \emph{preserving the original final answer}. This isolates process-reading from answer-checking: a model that only inspects the answer token will see no difference between $r_{\text{good}}$ and $r_{\text{bad}}$.
\end{itemize}

\section{Experiments}
\subsection{Experimental Setup}
\label{sec:tgs-setup}


\paragraph{Evaluated methods}
We evaluate calibration approaches spanning different training paradigms on two model scales.
As \emph{inference-stage} methods, we evaluate
Verbalize~\citep{can},
P(True)~\citep{language}, and
SteerConf~\citep{steerconf}.
As \emph{training-stage} methods, we consider
ConfTuner~\citep{conftuner} and
RLCR~\citep{beyond} as holistic-confidence baselines, and
VL-Calibration~\citep{vlcalibration} as a modality-aware baseline.
All training-stage methods share the same Qwen3-VL backbone and training data, isolating the effect of the calibration strategy.
For our main comparison, we report \textbf{Base} (Qwen3-VL-Instruct with verbalized prompting)
and \textbf{VL-Calibration}, the strongest method in terms of ECE,
on both 4B and 8B scales in Table~\ref{tab:merged_results}.

\begin{figure}[t]
    \centering
    \includegraphics[width=\columnwidth]{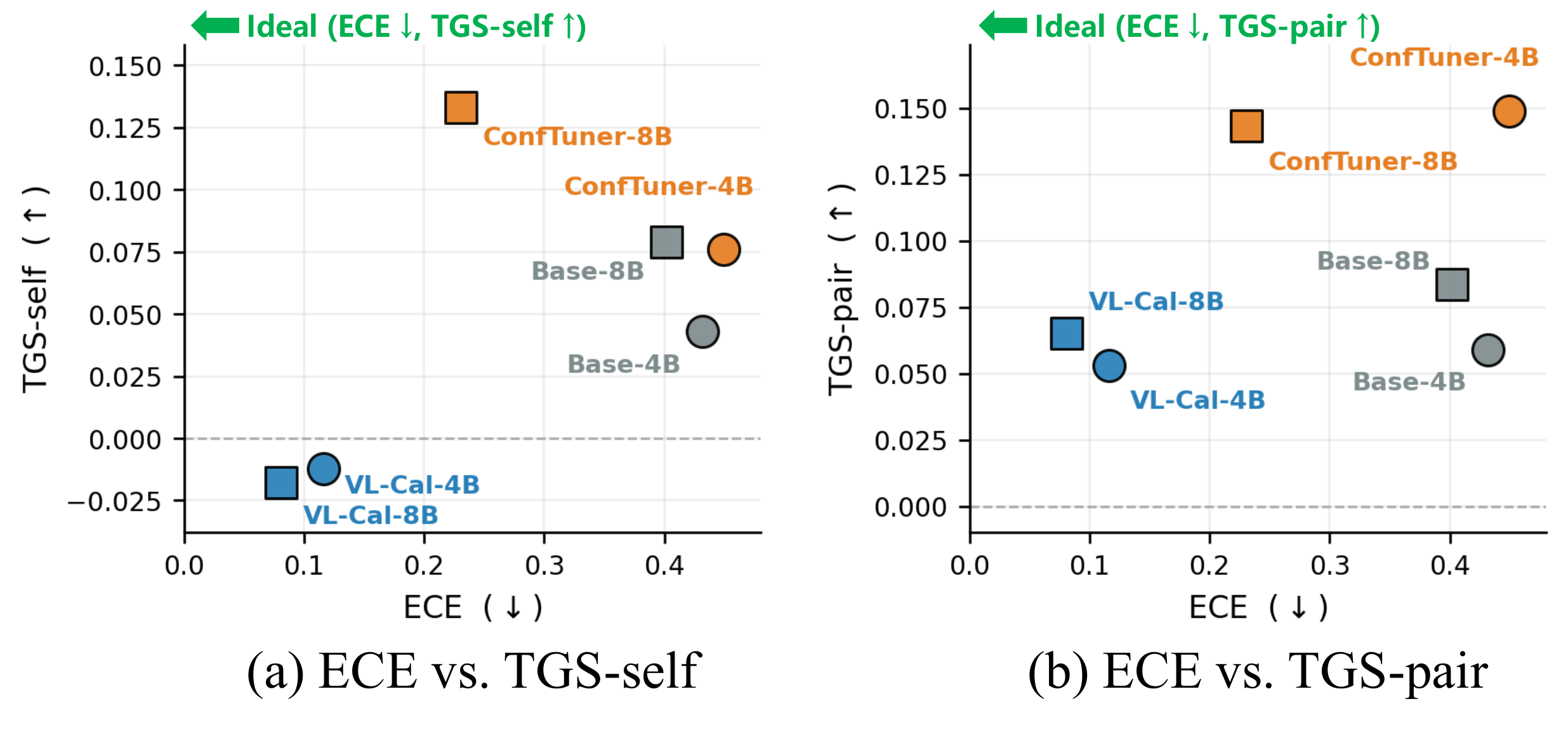}
    \caption{\textbf{Outcome-level calibration does not imply trajectory-grounding.}
    VL-Calibration moves models leftward (lower ECE) but not upward on either
    TGS-self (a) or TGS-pair (b), confirming that the disconnect holds for both
    self-trajectory reading and external trajectory discrimination.}
    \label{fig:ece_vs_tgs}
    \vspace{-0.5em}
\end{figure}

\paragraph{Metrics}
We report both outcome-level and trajectory-level metrics.
\emph{Outcome-level}: expected calibration error (ECE$\downarrow$) and area under the ROC curve (AUROC$\uparrow$).
\emph{Trajectory-level}: TGS-self$\uparrow$ and TGS-pair$\uparrow$ (averaged across vision, reasoning, and answer axes).

\paragraph{Benchmarks}
We evaluate on 10 benchmarks spanning mathematical and geometric reasoning
(DynaMath, Geo3K, MathVerse, MathVision, MathVista, WeMath),
logical reasoning (LogicVista),
and multi-discipline reasoning (A-OKVQA, MMK12, MMMU-Pro).

\begin{figure}[t]
  \centering
  \includegraphics[width=\columnwidth]{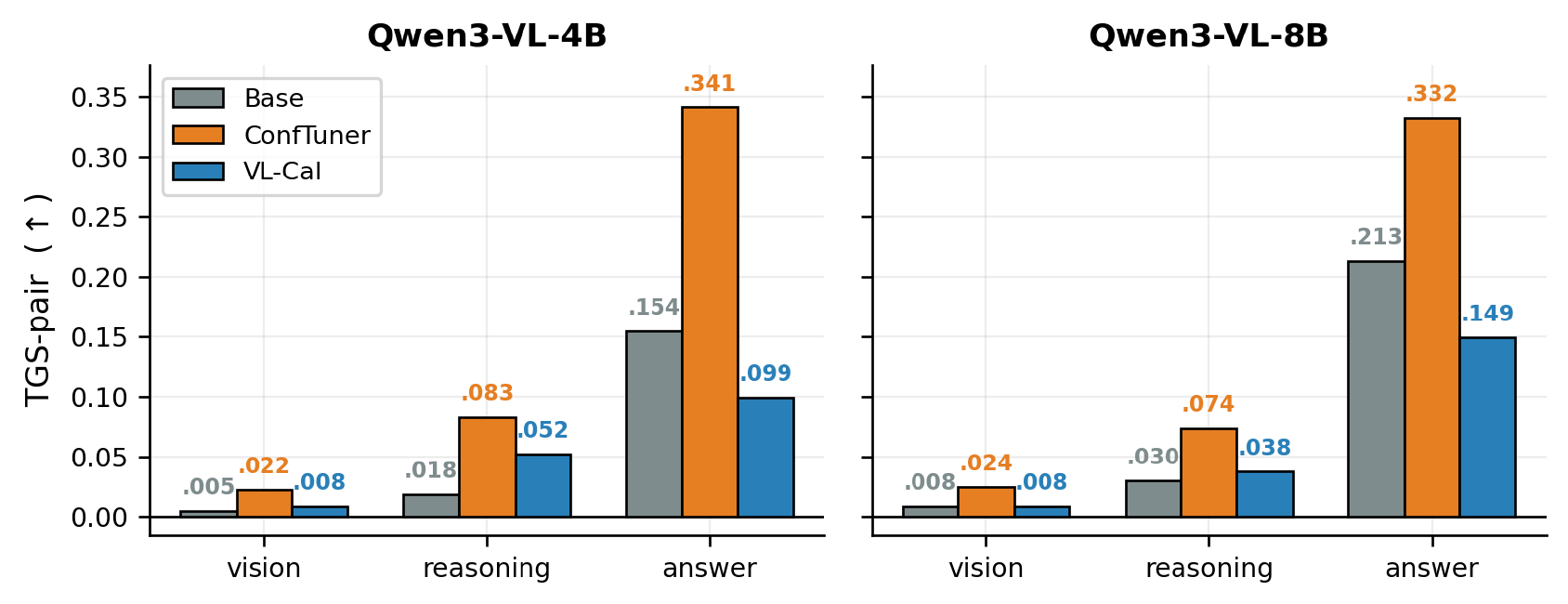}
  \caption{\textbf{TGS-pair by axis.}
Vision grounding is consistently the weakest axis across all methods and scales.
VL-Calibration shows low but positive TGS-pair on the vision axis at both scales,
and on the 8B scale, all methods show substantially lower scores on both vision and reasoning axes than on the answer axis, with grounding concentrated primarily in the answer axis.}
  \label{fig:tgs_axis}
\end{figure}

\begin{table}[t]
\centering
\small
\setlength{\tabcolsep}{3.2pt}
\begin{tabular}{@{}lcccc@{}}
\toprule
\textbf{Model}
& AUROC$\uparrow$
& ECE$\downarrow$
& TGS-self$\uparrow$
& TGS-pair$\uparrow$ \\
\midrule
Qwen3-VL-8B    & .664 & .275 & .063 & .084 \\
InternVL3.5-8B & .655 & .300 & .046 & .069 \\
\bottomrule
\end{tabular}
\caption{\textbf{Cross-architecture evaluation.}
Averages over 10 benchmarks; TGS-pair uses the same 18.3K TGS-Bench pairs for both models.}
\label{tab:cross_architecture}
\end{table}

\begin{table}[t]
\centering
\small
\setlength{\tabcolsep}{5pt}
\begin{tabular}{l cc cc}
\toprule
& \multicolumn{2}{c}{\textbf{Qwen3-VL-4B}}
& \multicolumn{2}{c}{\textbf{Qwen3-VL-8B}} \\
\cmidrule(lr){2-3} \cmidrule(lr){4-5}
\textbf{Generator} & Base & VL-Cal & Base & VL-Cal \\
\midrule
Qwen2.5-VL-72B & \textbf{.059} & \textbf{.053} & \textbf{.084} & \textbf{.065} \\
Qwen2.5-VL-32B & .057 & .040 & .081 & .043 \\
Qwen2.5-VL-7B  & .047 & .053 & .069 & .054 \\
\bottomrule
\end{tabular}
\caption{\textbf{Sensitivity of TGS-pair to generator model.}
TGS-pair is averaged across vision, reasoning, and answer axes and averaged across the 10 benchmarks on the same 5.3K matched records for all generators.}
\label{tab:generator_sensitivity}
\end{table}

\subsection{Results}
\label{sec:tgs-results}
Table~\ref{tab:merged_results} presents the main results. We highlight three observations.
\paragraph{Observation 1: Outcome-level gains do not transfer to trajectory-level grounding.}
VL-Calibration dramatically reduces ECE over the base model on both scales (4B: $.432 \to .116$; 8B: $.402 \to .081$), yet TGS-self remains near zero and turns negative after calibration, while TGS-pair decreases on both scales (4B: $.059 \to .053$; 8B: $.084 \to .065$).
This confirms that outcome-level calibration and trajectory-level grounding measure fundamentally different properties.
Figure~\ref{fig:ece_vs_tgs} visualizes this disconnect; VL-Calibration moves leftward (lower ECE) but not upward (higher TGS).
\paragraph{Observation 2: Vision grounding is the weakest axis.}
Per-axis analysis of TGS-pair (Figure~\ref{fig:tgs_axis}) reveals that the vision axis is consistently the weakest across all displayed methods and scales, indicating that models are least sensitive to the quality of their visual perception rationale.
This aligns with the trajectory-independence observed for the vision axis in \S\ref{sec:da} and suggests that vision-language grounding remains a critical bottleneck for confidence calibration.
\paragraph{Observation 3: Calibration training can induce anti-grounding.}
The signed formulation of TGS reveals cases where calibration training \emph{worsens} trajectory-grounding.
VL-Calibration yields negative TGS-self on 8 of 10 benchmarks at 4B and 9 of 10 at 8B, meaning that masking its own trajectory raises confidence on most benchmarks. In contrast, TGS-pair remains positive on every displayed benchmark.
Such self-trajectory anti-grounding is invisible to ECE but directly captured by TGS, underscoring the necessity of a trajectory-level diagnostic.
The same pattern extends to MM-Vet and GQA: VL-Calibration yields lower
TGS-self and TGS-pair than Base at both scales (Table~\ref{tab:broader_diagnostics}).


\paragraph{Cross-architecture check}
Table~\ref{tab:cross_architecture} shows weak TGS-self and TGS-pair for both Qwen3-VL-8B and InternVL3.5-8B. This pattern spans different vision encoders and multimodal architectures, but not independent language-model families, since InternVL3.5-8B uses a Qwen3-8B backbone.




\section{Analysis}
\subsection{Sensitivity to TGS-Bench Generator}
\label{sec:generator_sensitivity}
TGS-Bench uses Qwen2.5-VL-72B-Instruct as its default good-trajectory
generator. Inclusion is capability-gated, since an item enters the benchmark
only if the generator reaches the gold answer and completes the visual and
reasoning structure. Regenerating the 5.8K-item pool with smaller
generators retains 97.6\% of items at 32B and 93.4\% at 7B. A weaker
generator therefore drops items it cannot solve and biases the suite toward
easier ones, so we adopt the largest generator to minimize this selection
bias. To verify that our findings are not artifacts of this choice, we rebuild
the good trajectories with Qwen2.5-VL-32B and 7B and re-evaluate TGS-pair on
both scales.
As shown in Table~\ref{tab:generator_sensitivity}, TGS-pair values shift modestly across generators but remain below $.1$ in every setting, and VL-Calibration improves
over the base model in only one of the six generator--scale comparisons. 
Thus, the weak trajectory grounding is not driven by the size of the
good-trajectory generator.

\section{Conclusion}
We demonstrate that verbalized confidence is largely trajectory-independent in the Vision-Language Models and calibration methods we evaluate: it changes little when trajectory components are replaced, physically removed, or internally contradicted by the model's own self-doubt, and calibration training can paradoxically worsen this disconnect. 
To formalize trajectory-level evaluation, we propose the Trajectory-Grounding Score (TGS) in two complementary forms—TGS-self and TGS-pair—alongside TGS-Bench, a model-agnostic suite spanning 10 benchmarks with controlled good/bad trajectory pairs. Our experiments reveal that methods excelling on conventional metrics score near zero or negative on TGS, with vision grounding consistently the weakest axis, confirming that outcome-level calibration and trajectory-level grounding measure fundamentally different properties. We hope that TGS encourages future calibration methods that derive confidence from reasoning rather than from learned base rates.

\section*{Limitations}

Our evaluation covers the Qwen3-VL family (4B and 8B) with six calibration methods and includes a cross-architecture check on InternVL3.5-8B; extending the analysis to additional VLM architectures and larger scales would strengthen generalizability. The 10-benchmark suite predominantly consists of mathematical reasoning tasks with verifiable answers. Diagnostic subsets on MM-Vet and GQA broaden the task coverage, but generalization to unrestricted open-ended generation remains an open question. TGS-Bench relies on external models for trajectory construction; although our sensitivity analysis (Table~\ref{tab:generator_sensitivity}) shows robustness across generator scales, this dependency remains a limitation. Finally, this work is diagnostic in nature---we identify the problem but do not propose a training method to resolve it; developing calibration objectives that explicitly optimize for trajectory-grounding is an important direction for future work.

\section*{Ethics Statement}
This work diagnoses a reliability failure in VLMs where models appear well-calibrated under standard metrics while lacking genuine self-assessment, contributing to AI safety research. Our findings caution that current calibration metrics alone are insufficient for certifying confidence reliability in high-stakes deployment scenarios such as medical imaging or autonomous systems. All benchmarks used are publicly available; we will release TGS-Bench and evaluation code upon acceptance. We used AI writing assistants for language polishing; all technical content, experiments, and claims were verified by the authors.


\section*{Acknowledgments}

This work was partly supported by the 2026 Cultural Heritage Smart
Preservation and Utilization R\&D Program of Korea Heritage Service,
National Research Institute of Cultural Heritage (Project Name:
Development of AI Agent Technology for Architectural Heritage
Restoration Design, Project Number: RS-2026-25531766, Contribution
Rate: 50\%)
and partly supported by the Institute of Information and Communications Technology Planning and Evaluation (IITP) grant funded by the Korea Government (MSIT) [RS-2021-II211341,
Artificial Intelligence Graduate School Program (Chung-Ang University)].



\bibliography{custom}

\appendix
\section{Analysis Details}
\label{sec:appendix_A}

\subsection{Confidence Distribution Analysis}
\label{sec:appendix_A-1}

\subsubsection{Confidence elicitation templates}
All analyses in this section reuse the same prompting templates used for self-trajectory generation.
Below we provide the model-specific templates in abbreviated form.

\paragraph{Qwen3-VL prompt template}
For \textbf{Qwen3-VL-Instruct}, we use the following template to elicit a free-form trajectory and a holistic confidence score:
\begin{quote}
\small
\textbf{User Prompt}\\
\texttt{Question: \{question\}\{choices\_block\}}\\
\texttt{First, think step by step inside <think>...</think>. Inside the think block, briefly describe what you see in the image and how you derive the answer.}\\
\texttt{Then output your final answer in \textbackslash boxed\{...\} and a confidence score from 0 to 10 inside <confidence>...</confidence>.}\\
\texttt{Expected output format:}\\
\texttt{<think> ... </think>}\\
\texttt{\textbackslash boxed\{answer\}}\\
\texttt{<confidence>}\\
\texttt{[0--10]}\\
\texttt{</confidence>}
\end{quote}

\paragraph{VL-Calibration prompt template}
For \textbf{VL-Calibration}, we use the following template to elicit structured vision/reasoning trajectories and decoupled confidence scores:
\begin{quote}
\small
\textbf{System Prompt}\\
\texttt{You FIRST think through the reasoning process as an internal monologue, then provide the final answer.}\\
\texttt{The reasoning process MUST BE enclosed within <think> </think> tags. Inside the <think> tags, you MUST explicitly separate your thought process into two distinct parts: enclose your visual perception analysis within <vision> </vision> tags, and your logical deduction within <reasoning> </reasoning> tags. The final answer MUST BE put in \textbackslash boxed\{\}.}\\
\texttt{After that, perform an <analysis>...</analysis> block to analyse the vision and reasoning confidence in your answer.}\\
\texttt{Finally, output the confidence scores (0, 1, 2, 3, 4, 5, 6, 7, 8, 9, 10) enclosed within <confidence></confidence> tags. Inside the confidence tags, you MUST strictly output two separate scores enclosed within <vision\_confidence> </vision\_confidence> and <reasoning\_confidence> </reasoning\_confidence> tags respectively.}
\end{quote}

\begin{figure*}[t]
  \centering
  \includegraphics[width=0.8\textwidth]{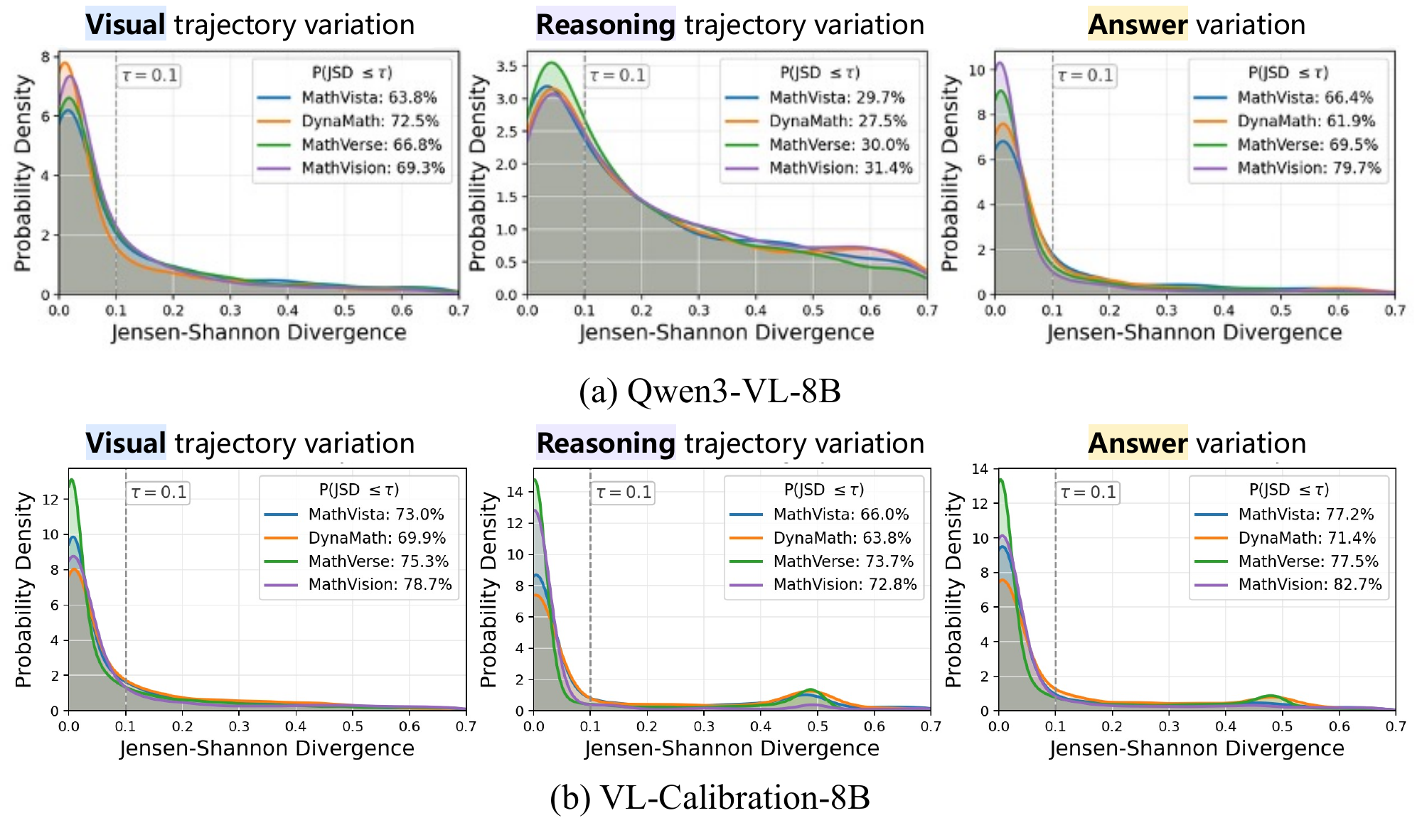}
  \caption{JSD distributions when varying trajectory components.
    \textbf{(a)} Qwen3-VL-8B (base). \textbf{(b)} VL-Calibration-8B.
    The base model is already substantially trajectory-independent
    (vision $P(\mathrm{JSD}\leq0.1)\approx64$--$73\%$).
    VL-Calibration increases this ratio
    (vision $P(\mathrm{JSD}\leq0.1)\approx70$--$79\%$),
    indicating that calibration training does not resolve the underlying
    disconnect between confidence and trajectory.}
  \label{fig:jsd_2}
\end{figure*}

\subsubsection{Trajectory construction}
\paragraph{Trajectory construction for Qwen3-VL}
Since Qwen3-VL-Instruct does not natively produce structured trajectory tags, we employ a two-stage procedure.
First, given an image $I$ and question $q$, we let the model generate a free-form chain-of-thought response.
Second, we prompt the model itself to segment the response into a \emph{vision} component $v$ (visual perception rationale) and a \emph{reasoning} component $r$ (logical/arithmetic chain), using the following structured extraction template:
\begin{quote}
\small
\textbf{Extraction Prompt}\\
\texttt{You are given a model response to a visual question. Split the response into two parts:}\\
\texttt{(1) <vision>: statements describing what is visually observed in the image, and}\\
\texttt{(2) <reasoning>: logical, arithmetic, or deductive steps used to derive the answer.}\\
\texttt{Do not add new information. Preserve the original meaning and wording as much as possible. Output only the two tagged fields.}
\end{quote}
We manually verified the segmentation quality on a random subset of 100 samples and found $>$95\% agreement with human annotation.

\paragraph{Trajectory construction for VL-Calibration}
VL-Calibration natively generates structured outputs with explicit \texttt{<vision>} and \texttt{<reasoning>} blocks, followed by separate confidence tokens $c_{\text{vis}}$ and $c_{\text{reas}}$.
We directly extract content within each tag pair without additional segmentation.

\subsubsection{Alternative trajectory generation}
For each sample, we generate $k=5$ alternative trajectories per axis using Qwen2.5-VL-7B-Instruct with diverse sampling (temperature $T=0.8$, top-$p=0.95$).
For the vision axis, we use the following template:
\begin{quote}
\small
\texttt{You are given an image, a question, and an existing answer. Describe the visual content relevant to the question in a different way from the original trajectory, while remaining consistent with the image and preserving the same final answer.}
\end{quote}
For the reasoning axis, we use:
\begin{quote}
\small
\texttt{You are given an image, a question, and an existing answer. Solve the problem using a different reasoning approach from the original trajectory, while preserving the same final answer.}
\end{quote}
For the answer axis, we pair the original trajectory with each of the other $k-1$ candidate answers.

\subsubsection{JSD computation}
For each pair of alternative trajectories $(x_i, x_j)$ along axis $x$, we elicit the full confidence distribution $P_{\mathcal{M}}(C=c)$ for $c \in \{0, 1, \ldots, 10\}$ from digit-token logits, then compute:
\[
\begin{aligned}
\mathrm{JSD}(P \| Q)
&= \tfrac{1}{2} D_{\mathrm{KL}}(P \| M)
 + \tfrac{1}{2} D_{\mathrm{KL}}(Q \| M), \\
M &= \tfrac{1}{2}(P + Q).
\end{aligned}
\]
For VL-Calibration, we compute JSD separately for $c_{\text{vis}}$ and $c_{\text{reas}}$ and report their average $\frac{1}{2}(\text{JSD}_{\text{vis}} + \text{JSD}_{\text{reas}})$ as the per-sample divergence.
All pairwise JSD values are computed across $\binom{k}{2}$ pairs per sample.

\subsubsection{Additional results across models and scales.}
\begin{figure*}[t]
    \centering
    \includegraphics[width=0.8\textwidth]{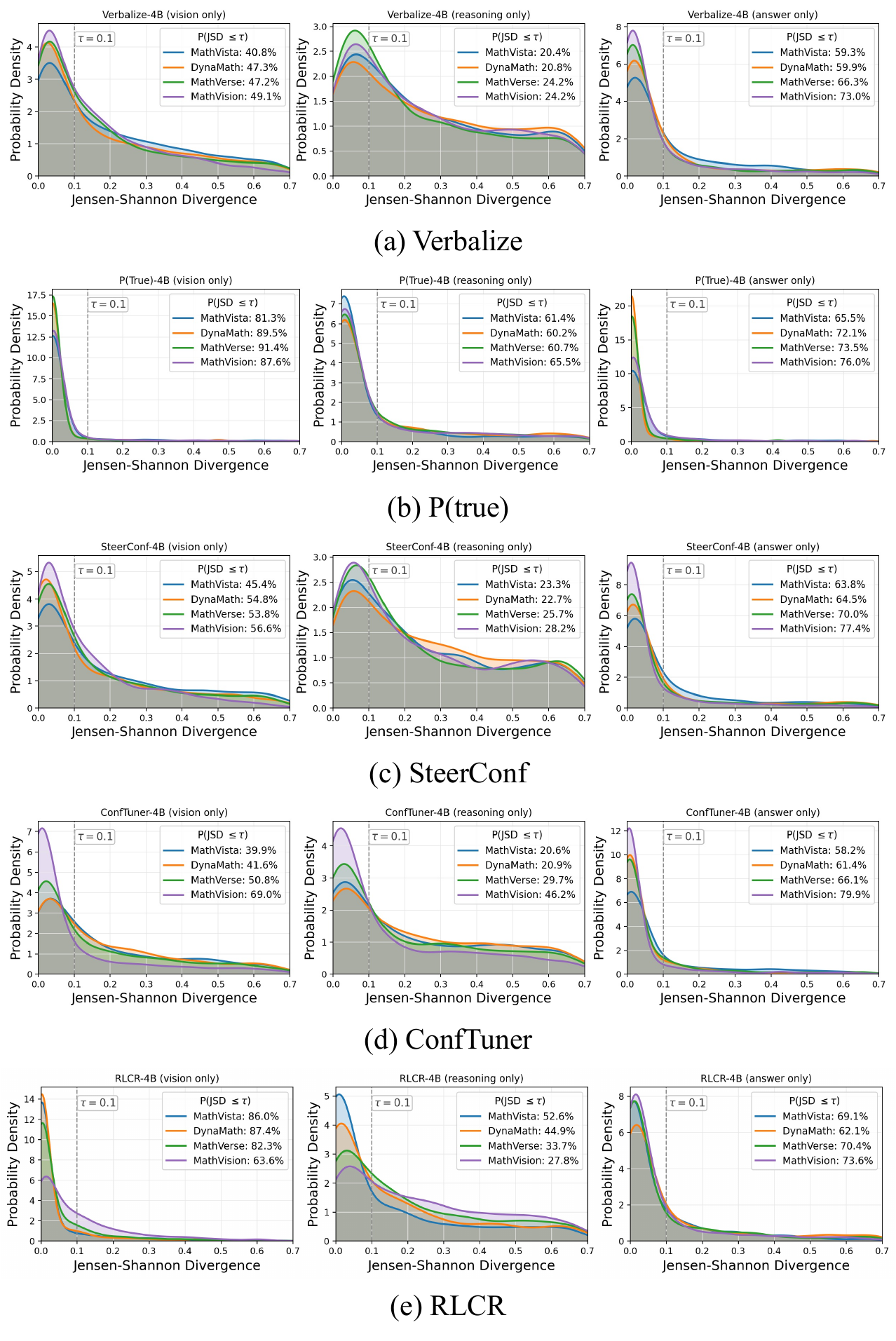}
    \caption{JSD distributions when varying trajectory components across \textbf{4B} confidence-estimation methods: \textbf{(a) Verbalize}, \textbf{(b) P(True)}, \textbf{(c) SteerConf}, \textbf{(d) ConfTuner}, and \textbf{(e) RLCR}. Across all 4B methods, the same qualitative pattern persists: confidence is most trajectory-independent on the vision axis, somewhat less so on the reasoning axis, and remains substantially insensitive to answer substitutions. Calibration training reduces but does not eliminate trajectory-independence.}
    \label{fig:jsd_4b_models}
\end{figure*}

\begin{figure*}[t]
    \centering
    \includegraphics[width=0.8\textwidth]{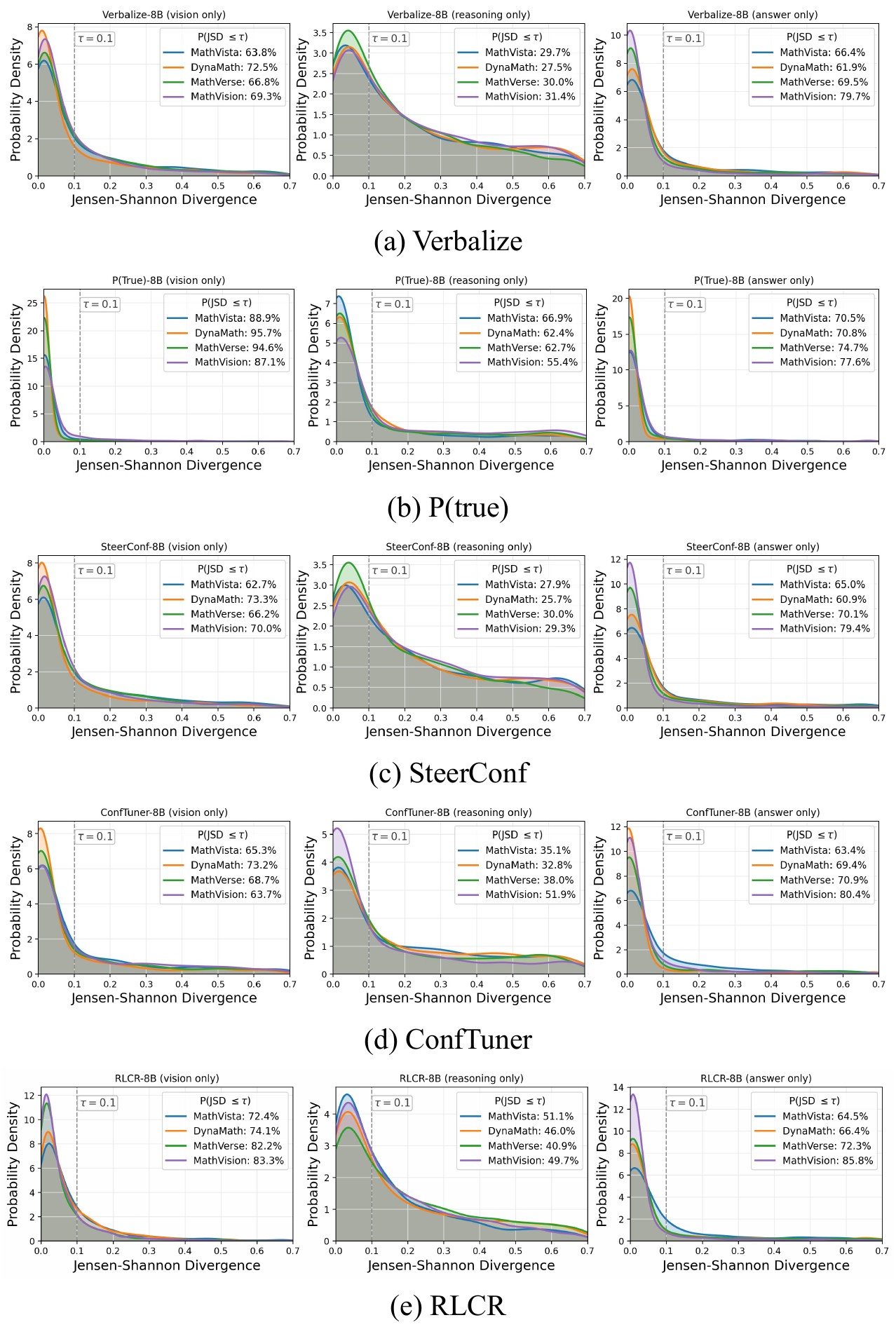}
    \caption{JSD distributions when varying trajectory components across \textbf{8B} confidence-estimation methods: \textbf{(a) Verbalize}, \textbf{(b) P(True)}, \textbf{(c) SteerConf}, \textbf{(d) ConfTuner}, and \textbf{(e) RLCR}. As in the 4B setting, confidence remains highly trajectory-independent, especially on the vision axis, with somewhat greater sensitivity on the reasoning axis. Calibration training weakens but does not resolve this independence.}
    \label{fig:jsd_8b_models}
\end{figure*}

The main text (Figure~\ref{fig:jsd}) reports JSD distributions for Qwen3-VL-4B and
VL-Calibration-4B. Figure~\ref{fig:jsd_2} reports the corresponding 8B
results for Qwen3-VL and VL-Calibration.
Figures~\ref{fig:jsd_4b_models} and \ref{fig:jsd_8b_models} extend the
analysis to Verbalize, P(True), SteerConf, ConfTuner, and RLCR at 4B and
8B, respectively. The core finding is consistent: trajectory-independence
holds across all models, with the vision axis showing the strongest
independence and the reasoning axis showing the weakest (though still
substantial).

\subsection{Token Masking Analysis}
\label{sec:appendix_A-2}

\paragraph{Attention-level masking}
Rather than deleting trajectory tokens from the input (which would shift positional encodings and introduce confounds), we implement masking at the attention level.
Specifically, for a trajectory spanning token positions $[t_{\text{start}}, t_{\text{end}}]$, we modify the causal attention mask so that all tokens at positions $t > t_{\text{end}}$ (i.e., the confidence-generation tokens) cannot attend to masked trajectory tokens.
The masked tokens remain in the sequence and retain their positional encodings, but are rendered invisible to downstream generation.

\paragraph{Masking ratios}
We apply masking at five nested ratios: 0\% (no masking), 25\%, 50\%, 75\%, and 100\% (full masking).
At each ratio $r$, we uniformly sample $\lfloor r \cdot (t_{\text{end}} - t_{\text{start}} + 1) \rfloor$ token positions within the trajectory span and mask them.
To ensure nested masking conditions, the masked subset at a lower ratio is always contained within the masked subset at a higher ratio for the same sample.
To reduce variance, we repeat each partial masking condition (25\%, 50\%, 75\%) with 3 different random seeds and report the average confidence.
The 0\% and 100\% conditions are deterministic and require no repetition.

\paragraph{Confidence extraction}
Under each masking condition, we let the model generate the confidence token(s) and record the expected confidence:
\[
\mu(C) = \sum_{c=0}^{10} \frac{c}{10} \cdot P(C=c),
\]
For $c\in\{0,\ldots,9\}$, $P(C=c)$ uses the corresponding single-token probability.
For $c=10$, we compute the joint probability
$P(\texttt{``1''})P(\texttt{``0''}\mid\texttt{``1''})$ with a second forward pass,
then normalize over all 11 candidates.
For VL-Calibration, we extract both $\mu(C_{\text{vis}})$ and $\mu(C_{\text{reas}})$ and report their harmonic mean as the holistic confidence.

\begin{figure*}[t]
    \centering
    \includegraphics[width=\textwidth]{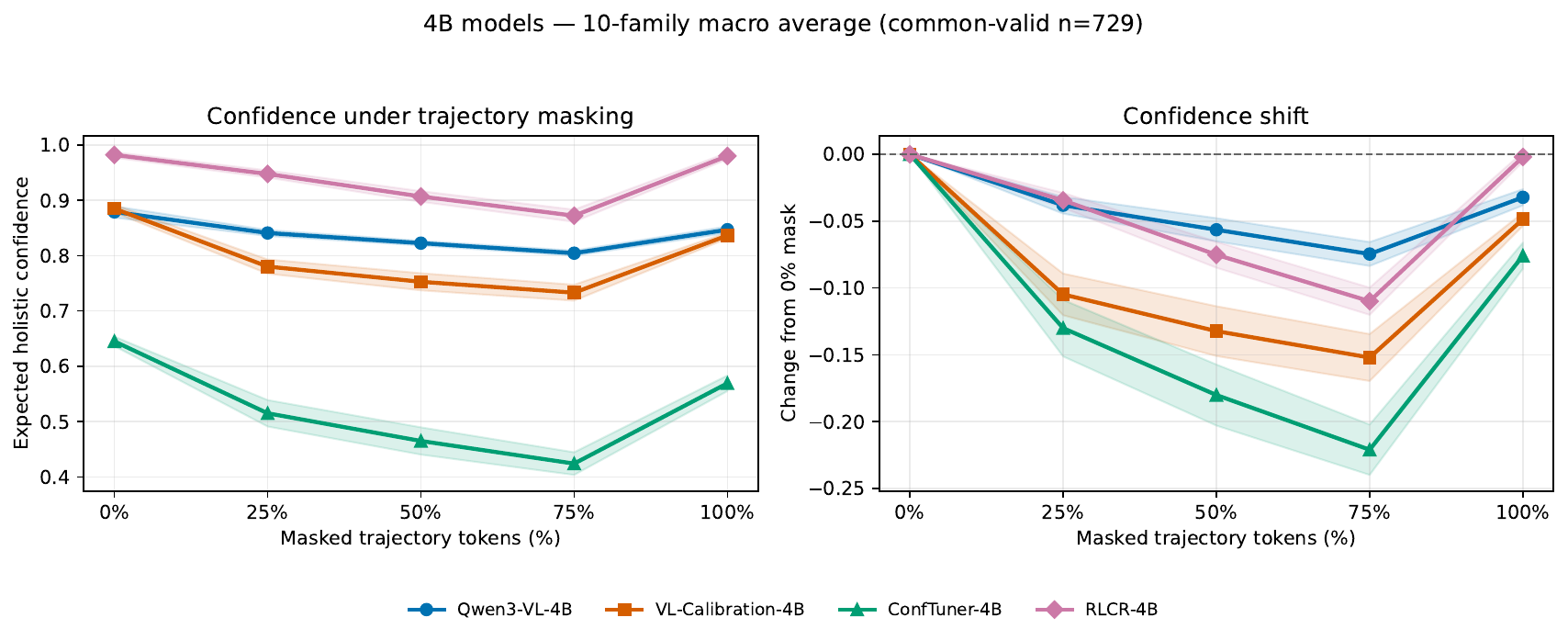}
    \caption{\textbf{Designated-rationale masking at 4B.}
    Expected holistic confidence under increasing masking for Base,
    VL-Calibration, ConfTuner, and RLCR. The left panel reports absolute
    confidence and the right panel its change from the unmasked condition.
    Curves are macro-averages across 10 benchmarks on the within-scale
    common-valid population ($n=729$); shading shows $\pm1$ standard error.}
    \label{fig:masking_4b_models}
\end{figure*}

\begin{figure*}[t]
    \centering
    \includegraphics[width=\textwidth]{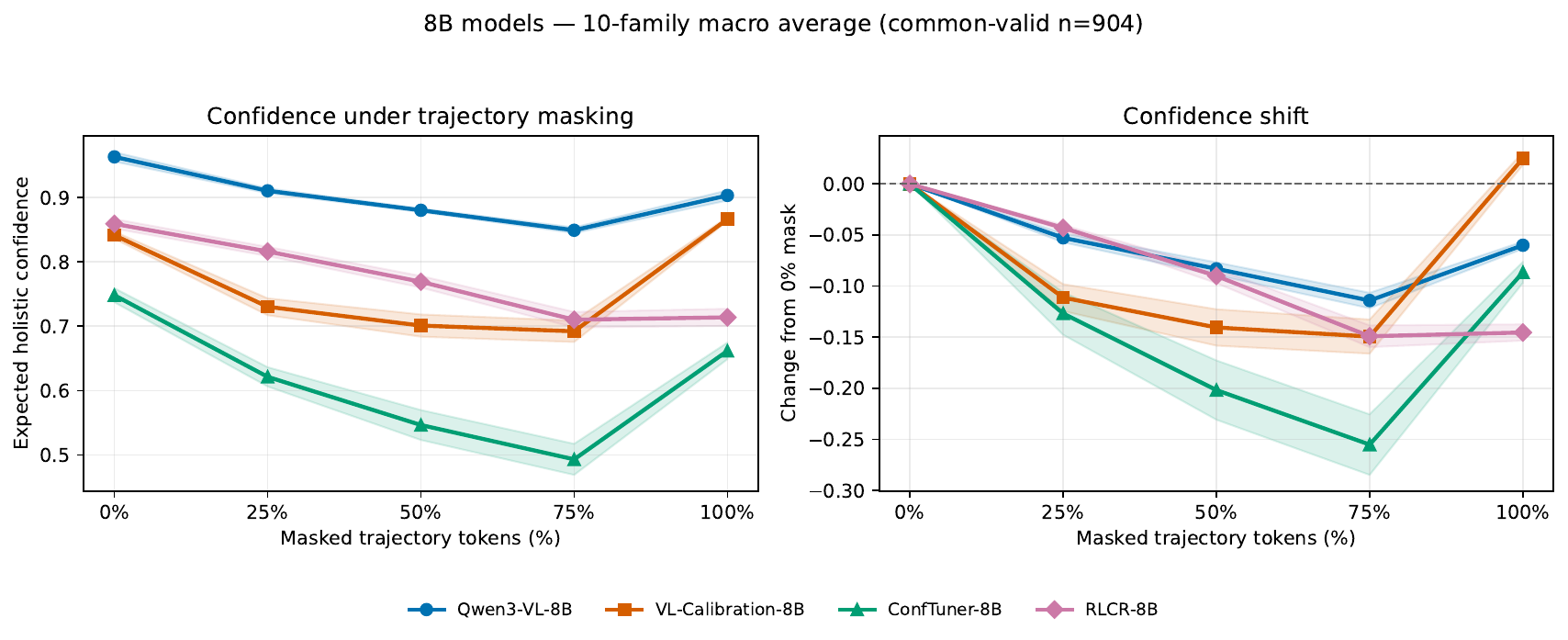}
    \caption{\textbf{Designated-rationale masking at 8B.}
    Expected holistic confidence under increasing masking for Base,
    VL-Calibration, ConfTuner, and RLCR. The left panel reports absolute
    confidence and the right panel its change from the unmasked condition.
    Curves are macro-averages across 10 benchmarks on the within-scale
    common-valid population ($n=904$); shading shows $\pm1$ standard error.}
    \label{fig:masking_8b_models}
\end{figure*}

\paragraph{Additional results across models and scales}
The main text (Figure~\ref{fig:masking}) reports masking curves for Qwen3-VL-4B and VL-Calibration-4B.
Figure~\ref{fig:masking_4b_models} and \ref{fig:masking_8b_models} consolidate the masking curves for all additional models and scales.
In all cases, the paradoxical uptick at 100\% masking persists, confirming that no evaluated model consistently relies on trajectory content when generating confidence.

\subsection{Intra-Trajectory Confidence Analysis}
\label{sec:appendix_A-3}

\paragraph{Hesitation marker lexicon}
We define hesitation markers as lexical cues indicating self-doubt or self-correction within the reasoning trajectory.
Our lexicon includes the following patterns (case-insensitive):
\begin{itemize}[nosep]
    \item \textit{Wait}, \textit{Hold on}, \textit{Hmm}
    \item \textit{Actually}, \textit{Let me reconsider}, \textit{On second thought}
    \item \textit{Let me recheck}, \textit{Let me recalculate}, \textit{Let me re-examine}
    \item \textit{I made a mistake}, \textit{I made an error}, \textit{That's not right}
    \item \textit{No, that's wrong}, \textit{Correction}, \textit{I should reconsider}
\end{itemize}
We match each pattern at word boundaries to avoid false positives
(e.g., ``wait'' within ``waiting'' is excluded).

\begin{table*}[t]
\centering
\small
\setlength{\tabcolsep}{4.5pt}
\begin{tabular}{@{}llccc ccc@{}}
\toprule
& & \multicolumn{3}{c}{\textbf{TGS-self}}
& \multicolumn{3}{c}{\textbf{TGS-pair}} \\
\cmidrule(lr){3-5} \cmidrule(lr){6-8}
\textbf{Scale} & \textbf{Model} & \textbf{Estimate [95\% CI]} & \textbf{SD} & $\boldsymbol{p}$
& \textbf{Estimate [95\% CI]} & \textbf{SD} & $\boldsymbol{p}$ \\
\midrule
4B & Base   & $.043~[.033,.053]$    & .017 & .002 & $.059~[.052,.066]$ & .012 & .002 \\
4B & VL-Cal & $-.012~[-.037,.013]$  & .041 & .322 & $.053~[.045,.063]$ & .015 & .002 \\
8B & Base   & $.079~[.065,.095]$    & .026 & .002 & $.084~[.076,.091]$ & .012 & .002 \\
8B & VL-Cal & $-.018~[-.029,-.005]$ & .020 & .049 & $.065~[.050,.078]$ & .024 & .002 \\
\bottomrule
\end{tabular}
\caption{\textbf{Statistical robustness of the main trajectory-level results.}}
\label{tab:tgs_robustness}
\end{table*}

\paragraph{Pre- and post-hesitation truncation}
For each hesitation marker with onset position $T_H$, we construct two truncated trajectories:
\begin{itemize}[nosep]
    \item \textbf{Pre-hesitation cut}: the trajectory ends immediately before the hesitation marker.
    \item \textbf{Post-hesitation cut}: the trajectory ends at the first sentence boundary following the marker, capped at the onset of the next hesitation marker when applicable.
\end{itemize}
For both conditions, we hold the original final answer fixed and elicit confidence using the exclusive 11-way digit readout over $\{0,\ldots,10\}$.
For Qwen3-VL, we use the holistic confidence score.
For VL-Calibration, we use the reasoning-confidence head for eligible reasoning-related hesitation markers.
This construction isolates the confidence change associated with the reconsideration segment while keeping the final answer unchanged.

\section{TGS-Bench Validation and Statistical Robustness}
\label{sec:tgs_validation}

\subsection{Construction and Independent-Judge Validation}
\label{sec:tgs_validation_construction}
The final TGS-Bench evaluation scores 18,295 records on each axis. Candidates
are textually distinct in 18,164 vision, 18,240 reasoning, and all 18,295
answer records; identical candidates contribute zero. Each comparison replaces
only its target axis, while the other two components are copied byte-for-byte
from the good trajectory. Vision and reasoning pairs have .958 character
similarity, change a median 1.6\% and 1.3\% of words, and preserve length.
Answer errors comprise multiple-choice letter
shifts (62.6\%), numeric or structured-numeric perturbations (36.6\%), and
Boolean flips (0.8\%).

For independent validation, image-aware InternVL3.5-8B judged a fixed,
benchmark-stratified sample of 400 pairs per axis with randomized A/B order.
Edited vision and reasoning trajectories retained high mean fluency (4.75 and
4.51 out of 5; 4.91 and 4.80 for good trajectories) and were judged natural in
93.0\% and 91.8\% of cases. The intended error side was identified in 75.5\%,
72.0\%, and 72.8\% of vision, reasoning, and answer pairs; edited answers were
judged plausible in 81.3\%. The judge classified 96.3\%, 90.8\%, and 67.8\% of
the respective edits as local. Under the corrected readout, confidence-change
correlations with edited-passage fluency are small for vision ($|r|\leq.10$)
but larger for reasoning ($|r|=.16$--$.22$), so the final data do not support a
uniform $|r|<.08$ bound.

Generator sensitivity is examined in \S\ref{sec:generator_sensitivity} and is
not repeated here.

\subsection{Statistical Robustness of the Main Results}
\label{sec:tgs_stats}
We recompute all statistics from the final corrected sample-level outputs.
Point estimates are averages across the 10 benchmarks.
We form 95\% percentile CIs with 5,000 two-stage bootstrap draws, resampling
benchmarks and then samples within each drawn benchmark, and report the standard deviation
across benchmark means. The $p$-values in Table~\ref{tab:tgs_robustness} use
exact two-sided Wilcoxon signed-rank tests of the ten benchmark means against
zero.

\section{Evaluation Details}
\label{sec:appendix_eval}

\subsection{Evaluation Metrics}

We use the following outcome-level and trajectory-level evaluation metrics:
\begin{enumerate}
    \item \textbf{Area Under the Receiver Operating Characteristic Curve (AUROC$\uparrow$):}
    Measures how well confidence distinguishes correct from incorrect answers across decision thresholds. Treating correctness as the positive label, we compute
    \begin{equation}
    \mathrm{AUROC}
    =
    \int_0^1
    \mathrm{TPR}\!\left(\mathrm{FPR}^{-1}(t)\right)\,dt,
    \end{equation}
    where TPR and FPR are the true-positive and false-positive rates.

    \item \textbf{Expected Calibration Error (ECE$\downarrow$):}
    Measures the discrepancy between empirical accuracy and confidence. We partition normalized confidence into $M=10$ equal-width bins $\{B_m\}_{m=1}^{M}$ on $[0,1]$ and compute
    \begin{equation}
    \mathrm{ECE}
    =
    \sum_{m=1}^{M}
    \frac{|B_m|}{N}
    \left|
    \operatorname{acc}(B_m)-\operatorname{conf}(B_m)
    \right|,
    \end{equation}
    where $N$ is the number of evaluated predictions, and $\operatorname{acc}(B_m)$ and $\operatorname{conf}(B_m)$ are the mean correctness and confidence in bin $B_m$.

    \item \textbf{Trajectory Grounding Score--Self (TGS-self$\uparrow$):}
    Measures the signed change in holistic confidence when the model's designated self-rationale is visible rather than fully attention-masked while the answer remains visible. A positive score indicates that confidence is grounded in the model's own trajectory (\S\ref{sec:tgs-self}).

    \item \textbf{Trajectory Grounding Score--Pair (TGS-pair$\uparrow$):}
    Measures the signed confidence difference between a good trajectory and its target-axis-perturbed counterpart. We compute it separately for the vision, reasoning, and answer axes and average the three scores (\S\ref{sec:tgs-pair}).
\end{enumerate}

\paragraph{Readout and aggregation}
Both TGS metrics use the corrected 11-way readout $\mu(C)=\sum_{c=0}^{10}(c/10)P(C{=}c)$. For VL-Calibration, the vision and reasoning axes use their corresponding heads, whereas the answer axis and TGS-self use the harmonic-mean confidence. Each intervention condition is evaluated once, and we average within benchmarks, then across the 10 benchmarks.

For outcome-level evaluation, Base uses its holistic verbalized confidence, while VL-Calibration uses the harmonic mean of its normalized visual and reasoning confidences. Their per-benchmark AUROC and ECE values are reproduced from the original VL-Calibration evaluation. ConfTuner ECE is computed locally from its checkpoint-native 0--9 digit distribution, $\sum_{k=0}^{9}(k/9)P(C{=}k)$, over valid completed outputs (73.9\% coverage at 4B and 76.1\% at 8B).

\subsection{Evaluation Datasets}

We evaluate on 10 benchmarks spanning mathematical and geometric reasoning, logical reasoning, and multi-discipline reasoning. We additionally evaluate diagnostic subsets of MM-Vet and GQA to assess transfer to broader visual reasoning tasks; these subsets are excluded from the 10-benchmark averages and reported separately in Table~\ref{tab:broader_diagnostics}.

\paragraph{Mathematical and geometric reasoning}
\begin{itemize}[leftmargin=*,itemsep=2pt,topsep=2pt]
    \item \textbf{DynaMath}~\citep{zou2025dynamath}: Uses programmatically generated variants of seed problems to test robust visual mathematical reasoning rather than memorization.
    \item \textbf{Geo3K}~\citep{lu2021inter}: Contains high-school geometry problems with dense formal-language annotations.
    \item \textbf{MathVerse}~\citep{zhang2024mathverse}: Provides multiple versions that progressively shift information from text to diagrams, probing genuine visual dependence.
    \item \textbf{MATH-Vision}~\citep{wang2024measuring}: Draws challenging problems from mathematical competitions across diverse subjects and difficulty levels.
    \item \textbf{MathVista}~\citep{lu2024mathvista}: Combines a broad range of mathematical and visual reasoning tasks.
    \item \textbf{WeMath}~\citep{qiao2025we}: Decomposes composite problems into subproblems organized by a hierarchy of mathematical concepts for fine-grained diagnosis.
\end{itemize}

\begin{table}[t]
\centering
\scriptsize
\setlength{\tabcolsep}{3.0pt}
\renewcommand{\arraystretch}{1.05}
\resizebox{\columnwidth}{!}{%
\begin{tabular}{@{}llcccc@{}}
\toprule
\textbf{Benchmark} & \textbf{Method}
& AUROC$\uparrow$ & ECE$\downarrow$
& TGS-self$\uparrow$ & TGS-pair$\uparrow$ \\
\midrule
\multicolumn{6}{@{}l}{\textbf{Qwen3-VL-4B}} \\
MM-Vet & Base   & .695 & .317 & .019 & .0465 \\
       & VL-Cal & .707 & .352 & $-$.003 & .0242 \\
GQA    & Base   & .624 & .200 & .010 & .1066 \\
       & VL-Cal & .614 & .223 & $-$.008 & .0497 \\
\midrule
\multicolumn{6}{@{}l}{\textbf{Qwen3-VL-8B}} \\
MM-Vet & Base   & .629 & .405 & .035 & .0534 \\
       & VL-Cal & .627 & .344 & $-$.011 & .0097 \\
GQA    & Base   & .610 & .338 & .062 & .1445 \\
       & VL-Cal & .600 & .246 & $-$.012 & .0105 \\
\bottomrule
\end{tabular}%
}
\caption{\textbf{Diagnostic transfer to broader visual reasoning tasks.}
Results on MM-Vet~\citep{yu2023mmvet} and GQA~\citep{hudson2019gqa}.}
\label{tab:broader_diagnostics}
\end{table}

\paragraph{Logical reasoning}
\begin{itemize}[leftmargin=*,itemsep=2pt,topsep=2pt]
    \item \textbf{LogicVista}~\citep{xiao2024logicvista}: Evaluates inductive, deductive, numerical, spatial, and mechanical reasoning across diverse visual formats.
\end{itemize}

\paragraph{Multi-discipline reasoning}
\begin{itemize}[leftmargin=*,itemsep=2pt,topsep=2pt]
    \item \textbf{A-OKVQA}~\citep{schwenk2022okvqa}: Requires commonsense and world knowledge beyond direct knowledge-base lookup.
    \item \textbf{MMK-12}~\citep{meng2025mm}: Covers K--12 multimodal STEM reasoning.
    \item \textbf{MMMU-Pro}~\citep{yue2025mmmu}: Strengthens MMMU by reducing text-only shortcuts, expanding answer choices, and introducing vision-only questions.
\end{itemize}

\begin{figure*}[t]
    \centering
    \includegraphics[width=\textwidth]{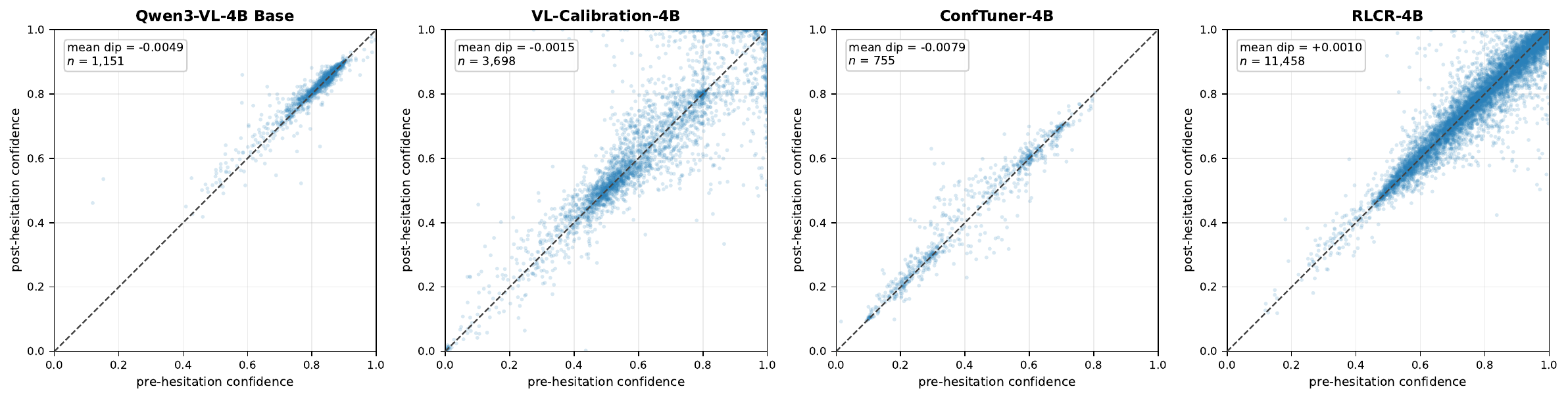}
    \caption{\textbf{Pre- versus post-hesitation confidence at 4B.}
    Each point is one eligible hesitation event for Base, VL-Calibration,
    ConfTuner, or RLCR. The dashed diagonal denotes no change; points below it
    have $c_{\mathrm{pre}}>c_{\mathrm{post}}$. Each panel reports the mean
    $\Delta c=c_{\mathrm{pre}}-c_{\mathrm{post}}$ and the number of scored events.}
    \label{fig:hesitation_4b_models}
\end{figure*}

\begin{figure*}[t]
    \centering
    \includegraphics[width=\textwidth]{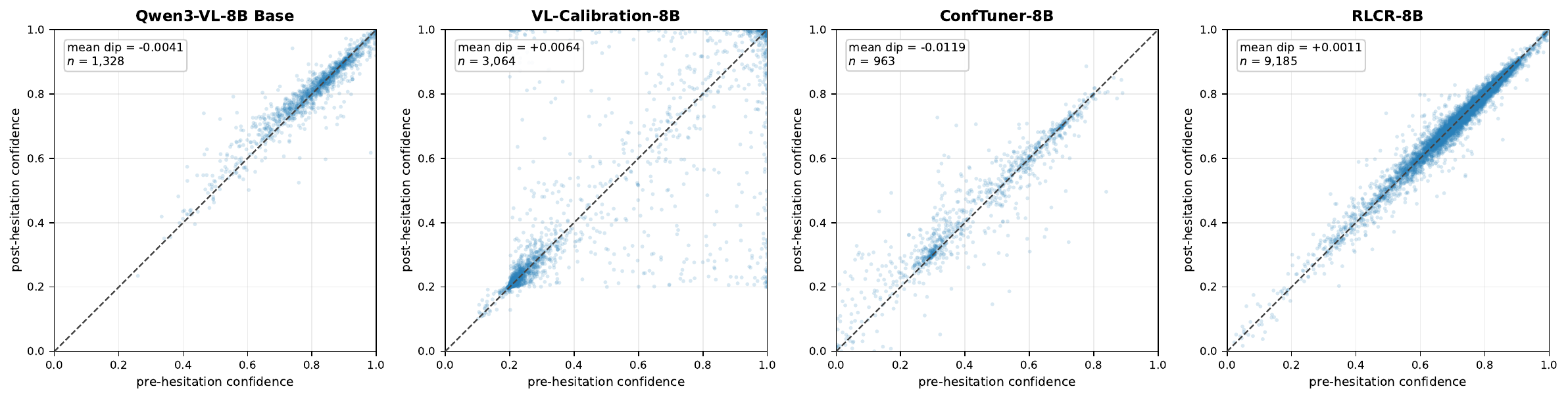}
    \caption{\textbf{Pre- versus post-hesitation confidence at 8B.}
    Each point is one eligible hesitation event for Base, VL-Calibration,
    ConfTuner, or RLCR. The dashed diagonal denotes no change; points below it
    have $c_{\mathrm{pre}}>c_{\mathrm{post}}$. Each panel reports the mean
    $\Delta c=c_{\mathrm{pre}}-c_{\mathrm{post}}$ and the number of scored events.}
    \label{fig:hesitation_8b_models}
\end{figure*}

\begin{figure*}[t]
    \centering
    \includegraphics[width=\textwidth]{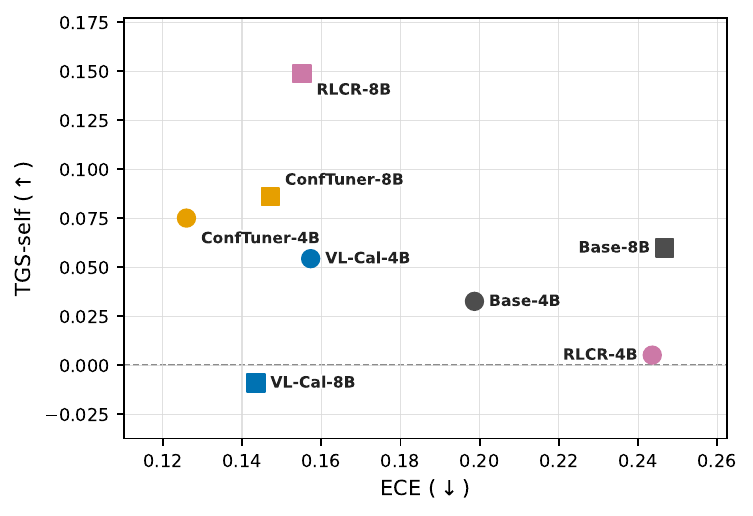}
    \caption{\textbf{Outcome calibration versus self-trajectory grounding.}
    Locally recomputed 10-bin ECE versus TGS-self for eight measured
    checkpoint--scale pairs: Base, VL-Calibration, ConfTuner, and RLCR at 4B and 8B.
    Both metrics use one trajectory per frozen instance and the panel-wide
    common-valid self population ($n=637$), followed by an 10 benchmark average. Lower ECE and higher TGS-self are preferred.}
    \label{fig:allmodels_ece_vs_tgs_self}
\end{figure*}

\begin{figure*}[t]
    \centering
    \includegraphics[width=\textwidth]{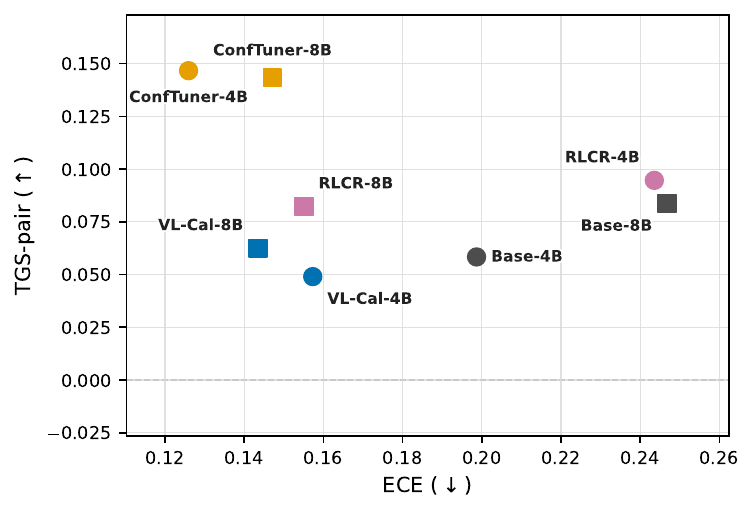}
    \caption{\textbf{Outcome calibration versus pairwise trajectory grounding.}
    Locally recomputed ECE versus TGS-pair for the same eight measured
    checkpoint--scale pairs. TGS-pair is shown as the raw signed confidence
    difference, not multiplied by $10^{-2}$. ECE and TGS-pair use their respective
    panel-wide common-valid self ($n=637$) and pair ($n=1{,}499$) populations and
    10 benchmark averages. Lower ECE and higher TGS-pair are preferred.}
    \label{fig:allmodels_ece_vs_tgs_pair}
\end{figure*}

\end{document}